\documentclass[runningheads]{llncs}

\usepackage{eccv}

\usepackage{eccvabbrv}

\usepackage{graphicx}
\usepackage{amssymb,mathrsfs,amsmath}
\usepackage{pythonhighlight}
\usepackage{tabularx}
\usepackage{multirow}
\usepackage{soul}
\usepackage{minitoc}
\usepackage{tocloft}
\usepackage{booktabs}
\usepackage{algorithm}
\usepackage{appendix}
\newenvironment{packeditemize}{
\begin{list}{$\bullet$}{
\setlength{\labelwidth}{8pt}
\setlength{\itemsep}{0pt}
\setlength{\leftmargin}{\labelwidth}
\addtolength{\leftmargin}{\labelsep}
\setlength{\parindent}{0pt}
\setlength{\listparindent}{\parindent}
\setlength{\parsep}{0pt}
\setlength{\topsep}{3pt}}}{\end{list}}
\definecolor{mygray}{HTML}{d4d4d4}
\definecolor{mygreen}{HTML}{cbedbd}

\usepackage[accsupp]{axessibility}  

\usepackage[pagebackref,breaklinks,colorlinks,citecolor=eccvblue]{hyperref}

\usepackage{orcidlink}

\begin{document}
\doparttoc 
\faketableofcontents 

\newcommand{\AlgName}{\textsc{JigMark}}
\newcommand{\SimName}{\textsc{HAV}}

\newcommand\DoToC{%
  \startcontents
  \printcontents{}{2}{\textbf{Contents}\vskip3pt\hrule\vskip5pt}
  \vskip3pt\hrule\vskip5pt
}

\newcommand*{\graybox}[1]{{%
    \ttfamily
    \hyphenchar\font=45\relax
    \setlength{\fboxsep}{4pt}
    \colorbox{mygray}{\makebox[6pt][l]{#1}}
}}

\newcommand*{\grayboxl}[1]{{%
    \ttfamily
    \hyphenchar\font=45\relax
    \setlength{\fboxsep}{4pt}
    \colorbox{mygray}{\makebox[8pt][l]{#1}}
}}

\newcommand*{\grayboxlt}[1]{{%
    \ttfamily
    \hyphenchar\font=45\relax
    \setlength{\fboxsep}{4pt}
    \colorbox{mygray}{\makebox[12pt][l]{#1}}
}}

\newcommand*{\grayboxt}[1]{{%
    \ttfamily
    \hyphenchar\font=45\relax
    \setlength{\fboxsep}{2pt}
    \colorbox{mygray}{\makebox[8pt][l]{#1}}
}}

\newcommand*{\grayboxltx}[1]{{%
    \ttfamily
    \hyphenchar\font=45\relax
    \setlength{\fboxsep}{1pt}
    \colorbox{mygray}{\makebox[12pt][l]{#1}}
}}

\newcommand*{\grayboxlll}[1]{{%
    \ttfamily
    \hyphenchar\font=45\relax
    \setlength{\fboxsep}{2pt}
    \colorbox{mygray}{\makebox[50pt][l]{#1}}
}}

\newcommand*{\greenbox}[1]{{%
    \ttfamily
    \hyphenchar\font=45\relax
    \setlength{\fboxsep}{4pt}
    \colorbox{mygreen}{\makebox[8pt][l]{#1}}
}}

\newcommand*{\greenboxl}[1]{{%
    \ttfamily
    \hyphenchar\font=45\relax
    \setlength{\fboxsep}{3pt}
    \colorbox{mygreen}{\makebox[12pt][l]{#1}}
}}

\newcommand*{\greenboxtl}[1]{{%
    \ttfamily
    \hyphenchar\font=45\relax
    \setlength{\fboxsep}{1pt}
    \colorbox{mygreen}{\makebox[12pt][l]{#1}}
}}
\newcommand{\samelineand}{\qquad}
\renewcommand{\thanks}[1]{\footnotetext[0]{#1}}

\lstnewenvironment{PythonB}[1][]{
  \lstset{style=mypython, frame=none, #1}
}{}

\title{\AlgName: A Black-Box Approach for Enhancing Image Watermarks against Diffusion Model Edits} 

\titlerunning{JIGMARK}


\author{Minzhou Pan\inst{1}*\thanks{*Equal contribution.}\and
Yi Zeng\inst{2}* \and
Ning Yu\inst{3} \and 
Cho-Jui Hsieh\inst{4} \and
Peter Henderson\inst{5} \and
Ruoxi Jia\inst{2} \and
Xue Lin\inst{1}
}

\authorrunning{M.~Pan et al.}


\institute{$^{1}$Northeastern University \samelineand $^{2}$Virginia Tech \samelineand $^{3}$Netflix Eyeline Studios \\ $^{4}$University of California,
Los Angeles \samelineand $^{5}$Princeton University}

\maketitle

\begin{abstract}
In this study, we investigate the vulnerability of image watermarks to diffusion-model-based image editing, a challenge exacerbated by the computational cost of accessing gradient information and the closed-source nature of many diffusion models. To address this issue, we introduce $\AlgName$. This first-of-its-kind watermarking technique enhances robustness through contrastive learning with pairs of images, processed and unprocessed by diffusion models, without needing a direct backpropagation of the diffusion process. Our evaluation reveals that $\AlgName$ significantly surpasses existing watermarking solutions in resilience to diffusion-model edits, demonstrating a True Positive Rate more than triple that of leading baselines at a 1\% False Positive Rate while preserving image quality. At the same time, it consistently improves the robustness against other conventional perturbations (like JPEG, blurring, etc.) and malicious watermark attacks over the state-of-the-art, often by a large margin. Furthermore, we propose the Human Aligned Variation ($\SimName$) score, a new metric that surpasses traditional similarity measures in quantifying the amount of image derivatives from image editing. The source code for this project is available on  \href{https://github.com/pmzzs/JigMark}{here}.

\end{abstract}

\section{Introduction}
\label{sec:intro}

\begin{figure}[h!]
    \begin{center}
    \includegraphics[width=1.0\linewidth]{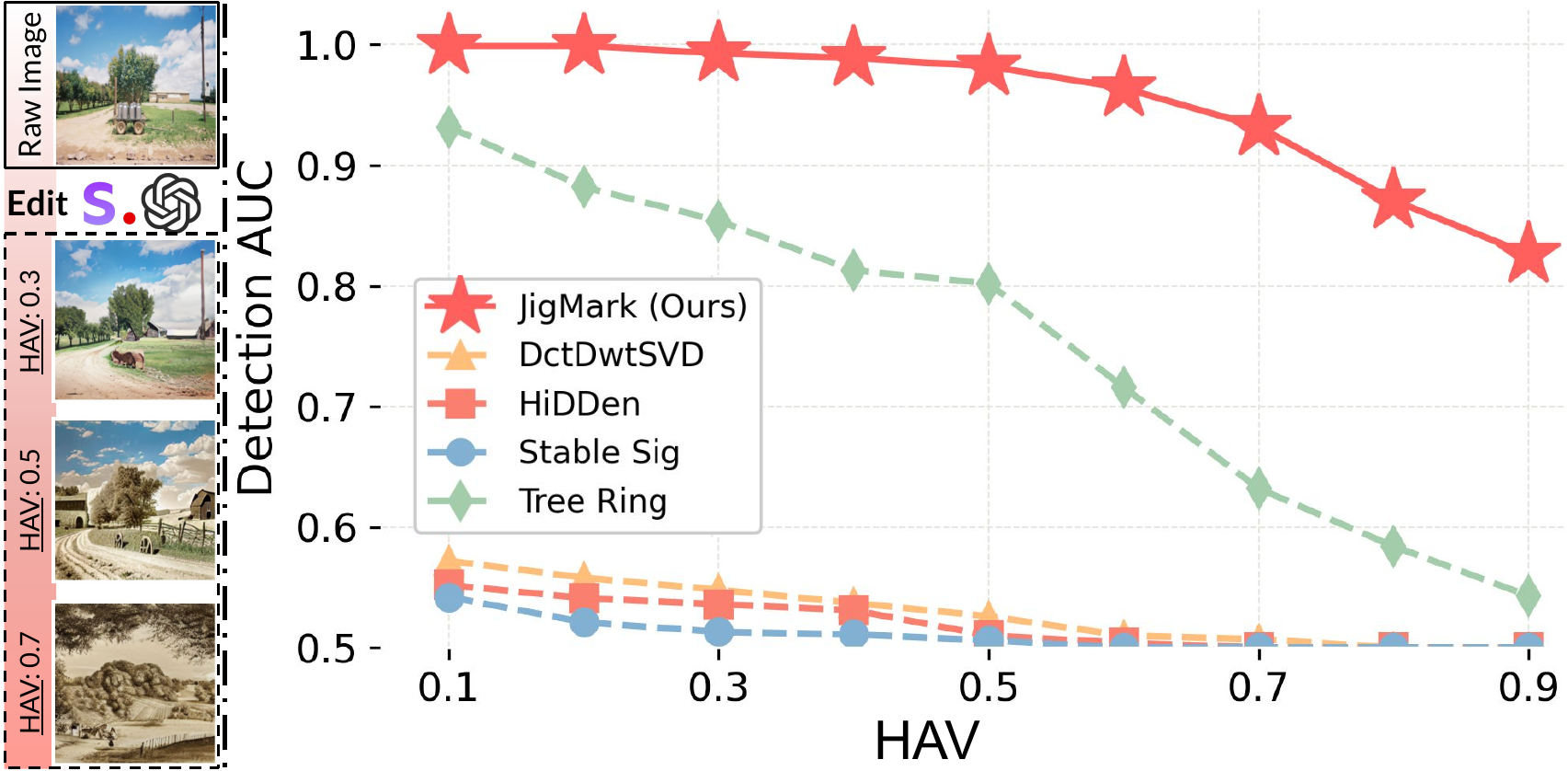}
    \end{center}
    \vspace{-1.5em}
     \caption{\textbf{Left}: image sample from different HAV values. As the $\SimName$ gets higher, the image becomes more dissimilar. \textbf{Right}: The trade-off between editingand detectability. As \textit{Human Aligned Variations} ($\SimName$) increase, $\AlgName$ maintains higher detection AUC than baseline methods~\cite{4554423, zhu2018hidden, zhu2018hidden, fernandez2023stable}.}
    \label{fig:tracerplot}
    \vspace{-2.0em}
\end{figure}

Diffusion models, such as Stable Diffusion~\cite{rombach2022high} and DALL·E 2~\cite{ramesh2022hierarchical}, have revolutionized image editing by enabling users to produce high-quality derived versions of image contents effortlessly. These models can perform complex operations, including object addition, removal, and style transfer, and have been incorporated in mainstream image editing tools like Adobe Photoshop~\cite{adobeai} and Google Photos~\cite{googleai}, reaching billions of users. However, the nature of noise addition and removal during the diffusion-based editing process can significantly impair the detectability of embedded watermarks in the edited watermarked images, as illustrated in Fig.~\ref{fig:tracerplot}. This poses a severe threat to the integrity of watermarking systems, undermining their effectiveness in protecting intellectual property (IP) rights and ensuring image authenticity.

Despite advancements in invisible image watermarking techniques~\cite{4554423, zhu2018hidden, wen2023tree, fernandez2023stable},
the surge in diffusion-model-based image editing presents new technical challenges impeding addressing the threat.
Traditional frequency domain watermarking methods~\cite{4554423} are ineffective against diffusion-model-induced perturbations and cannot improve themselves with information about potential downstream perturbations into their design process to enhance the robustness. While some deep-learning watermarking techniques~\cite{zhu2018hidden} are theoretically adaptable to new perturbations, such adaptation relies on the gradient of the target perturbation. The computational intensity of backpropagating through the diffusion process severely limits their applicability. 
Not to mention the prevalence of closed-source diffusion models in popular editing tools further obstructs the access to the gradient information.

To tackle these challenges, we introduce $\AlgName$, a first-of-its-kind watermarking method that gradually acquires robustness through contrastively learning from images with and without diffusion perturbations. Unlike previous learning-based watermarking methods that rely on continuous differentiable computational paths, $\AlgName$ requires only the non-modified original image and the diffusion-generated results as pairs. This approach ensures adaptability to overcome the computationally expensive direct backpropagation process or the inaccessibility of the computational path of close source models.

To support our evaluation and better understand the impact of diffusion model perturbations, we also propose the Human Aligned Variation ($\SimName$) score. This human-centric metric more accurately reflects the information derivatives as perceived by humans compared to conventional image similarity metrics (e.g., MSE, SSIM~\cite{wang2004image}, LPIPS~\cite{zhang2018unreasonable}). $\SimName$ also accurately reflects the strength of diffusion model perturbations and helps standardize efficacy comparisons between different watermarking techniques.
Our main contributions are as follows:

\begin{packeditemize}
\item \textbf{(i)} Revealing the vulnerability of existing watermarks to diffusion model editing, emphasizing the need for diffusion-resilient watermarking;
\item \textbf{(ii)} Proposing $\AlgName$, a black-box adaptable and robust watermarking technique grounds in contrastive learning without requiring direct access to perturbations' computational path;
\item \textbf{(iii)} Introducing the $\SimName$ score, a supporting metric for assessing image variations caused by diffusion models and enabling fairer comparisons;
\item \textbf{(vi)} Conducting extensive design analysis, including loss functions, model structure, and training methods;
\item \textbf{(v)} Performing comprehensive evaluations demonstrating $\AlgName$'s robustness against diffusion-model-based image editing, as well as its consistent improvements in robustness against conventional perturbations and watermark removal attacks over state-of-the-art methods.
\end{packeditemize}

\vspace{-1.5em}

\section{Preliminaries}

\subsection{Image Editing with Diffusion Models}
The evolution of image generation technologies has reached a significant milestone with the development of diffusion models~\cite{dhariwal2021diffusion}. In contrast to traditional generative adversarial network (GAN)-based methods~\cite{isola2017image, zhu2017unpaired} that make modifications in a single step, diffusion models adopt a novel approach by iteratively adding and removing noise in a multi-step process, leading to the achievement of photorealistic outcomes~\cite{dhariwal2021diffusion, meng2021sdedit, brooks2023instructpix2pix}. Such advancements have enabled a series of image editing tasks like text-guided image editing~\cite{meng2021sdedit}, image editing with instruction~\cite{brooks2023instructpix2pix}, perspective generation~\cite{liu2023zero}, and inpainting \cite{lugmayr2022repaint}.

The increasing accessibility of diffusion model-based image editing tools~\cite{adobeai, googleai} has lowered the barrier for users to modify existing images, potentially infringing upon content creators' rights. Moreover, the complex, multi-step process employed by diffusion models can inadvertently destroy or corrupt embedded watermarks, making it difficult for rights holders to track and enforce their IP rights. The combination of increased ease in creating unauthorized derivatives and the failure of existing watermarking methods to withstand diffusion model-based editing highlights the urgent need for more robust watermarking techniques. As diffusion models continue to advance and become more widely accessible, it is crucial to develop watermarking methods that can resist the perturbations introduced by these models, ensuring that content creators can effectively protect their IP in the face of these new challenges.


\subsection{Revisiting Existing Image Watermarks}
Unfortunately, no existing watermarking method is immune to the modifications introduced by the diffusion model.
Traditional image watermarking employs hand-crafted keys embedded in the frequency domain~\cite{465537, o1997rotation}. Effective under conventional conditions, these methods, however, struggle with novel distortions introduced by advanced diffusion models.
Recent advancements in deep learning, particularly with encoder-decoder architectures in watermarking~\cite{zhu2018hidden, tancik2020stegastamp}, have opened up new possibilities. These methods rely on a gradient path between the encoder and decoder for effective watermark learning. However, when faced with non-differentiable perturbations, typical approaches involve employing differentiable substitutes~\cite{shin2017jpeg} or integrating task-specific networks~\cite{zhang2021towards, fang2023flow}. Unfortunately, given the complexity of diffusion models, finding such differentiable substitutes is currently infeasible.

Another line of work has focused on integrating watermarks into the generative processes of diffusion models \cite{liu2023watermarking, ma2023generative, fernandez2023stable, wen2023tree}. However, these methods remain rooted in traditional frameworks. For example, Stable Signature \cite{fernandez2023stable} adapts HiDDeN's decoder \cite{zhu2018hidden} for diffusion models but demonstrates less robustness (see Fig.~\ref{fig:tracerplot}), and its robustness still relies on the gradient of perturbation. On the other hand, Tree-ring \cite{wen2023tree} embeds hand-crafted keys in the latent space of diffusion models as watermarks. Since the latent space is less susceptible to perturbations, Tree-ring exhibits better perturbation resistance. However, because the trigger generation and embedding process is not optimizable, it still lacks adaptability to unseen perturbations or any perturbation stronger than its assumptions, such as diffusion-based perturbations. Moreover, these watermarks are specifically designed for the outputs of their respective diffusion models, lacking the flexibility to provide IP protection for any given image, particularly real-world images or paintings produced by humans.

\subsection{Watermark Detectability}
Recent studies in text-based watermarking of AI-generated content \cite{zhang2023watermarks,sadasivan2023can} 
suggest that text-based watermarks of AI-generated content can be removed by quality-preserving perturbations. These perturbations result in high-quality contents as measured by distribution distance from a human reference distribution \cite{sadasivan2023can} or through a ``quality oracle'' \cite{zhang2023watermarks}.  However, it is important to note that a quality-preserving perturbation could significantly alter the watermarked content and the resulting material may not necessarily retain similarity to the original one. In contrast, motivated by IP protection,
our goal is to ensure the robustness of watermarks against those perturbations that still preserve similarity between the original and derived image---specifically, those perturbations that would be considered an IP violation under existing regulations~\cite{uscode2023}, rather than focusing on robustness against any arbitrary quality-preserving perturbation.
 Hence, their results~\cite{zhang2023watermarks,sadasivan2023can} regarding the impossibility of watermarking are not directly applicable to our context. 

\section{Threat Model}
\label{sec:threatmodels}

This section will provide an overview of the threat model considered in this paper.


\noindent
\textbf{Types of Perturbations.}
We categorize the perturbations that watermarked images may encounter into three types:

\begin{packeditemize}
\item \textbf{Type 1, Conventional Perturbations:} This category encompasses common perturbations such as JPEG compression, image blurring, mirroring, and rotation, which are frequently encountered during standard image transmission or exchange over the internet. Prior watermarking techniques~\cite{4554423, zhu2018hidden, wen2023tree, fernandez2023stable} have extensively addressed this type of perturbation.

\item \textbf{Type 2, Diffusion Perturbations:} We introduce this novel type of perturbation to address the emerging threat posed by the advent of diffusion models~\cite{meng2021sdedit, brooks2023instructpix2pix}. In this category, \emph{the editing strength should be reasonably constrained to ensure that the perturbed image remains semantically similar to the original while still generating meaningful modifications}. To better quantify the acceptable range of editing strengths, we propose $\SimName$ in Section~\ref{sec:hav}.

\item \textbf{Type 3, Watermark Removal Attacks:} These attacks are specifically designed to remove watermarks while preserving image quality. Attackers~\cite{zhao2023invisible, jiang2023evading, saberi2023robustness, lukas2023leveraging} often employ black-box attacks on watermarks using only a limited set of watermarked samples, making this type of attack a significant threat to watermark robustness.
\end{packeditemize}

\noindent
\textbf{Knowledge of the Watermark Agent.}
The watermark agent is tasked with embedding watermarks that remain detectable through various perturbations. 

\begin{packeditemize}
\vspace{-.5em}
\item \textbf{Dataset Familiarity:} The watermarking agent has access to and knowledge of the datasets containing the images to be watermarked. The watermarking agent can obtain and watermark any image within these datasets.

\item \textbf{Potential Perturbations:} The agent operates under the realistic constraint of limited knowledge about perturbation mechanisms, particularly considering closed-source systems like DALL·E 2~\cite{ramesh2022hierarchical}. The agent can only access pairs of original and perturbed image samples, differing from methods that assume explicit knowledge of perturbation mechanisms~\cite{zhu2018hidden, fernandez2023stable}.

\vspace{-.5em}
\end{packeditemize}

\noindent
\textbf{Evaluation and Validation:} We categorize the key aspects of watermark evaluation as follows:

\begin{packeditemize}
\vspace{-.5em}
\item \textbf{Perturbation Scope:} Watermarked images should be tested against both anticipated perturbations and unseen perturbations, such as those introduced by diffusion models that are not included in the agent's knowledge base. The considered perturbations should be capable of causing meaningful changes to the image while preserving perceptual similarity to the original content from a human perspective.

\item \textbf{Visual Stealthiness:} Essential for watermarking is the invisible, inconspicuous embedding of watermarks, which guarantees minimal visual disturbance while preserving the watermark's detectability.

\item \textbf{Adaptability to Personalized Keys:} A watermarking system should be able to adapt to new personalized keys to distinguish the different IP holders. Successful implementations, as seen in using random keys \cite{465537, o1997rotation,wen2023tree}, demonstrate this flexibility. This approach is notably more practical than systems requiring training new encoder-decoder pairs for each IP holder \cite{fernandez2023stable}.

\item \textbf{False Positive Rate Evaluation:} A comprehensive false-positive analysis is crucial, evaluating both non-watermarked images and scenarios where different IP holders use the same watermarking system with personalized keys. Testing for mistaken classification ensures the system's robustness across various users.

\vspace{-.5em}
\end{packeditemize}

\section{\AlgName: Our Approach}
\label{sec:methodology}
Building upon the threat models, $\AlgName$ innovates to develop an \textbf{optimizable watermarking system} that is \textbf{generalizable} to incorporate \textbf{black-box perturbations} without requiring explicit backpropagations while also being \textbf{customizable} for the integration of personalized keys for different IP holders without retraining the system again from scratch.

\subsection{Towards the black-box optimizable watermark}

\noindent
\textbf{Adapting Contrastive Learning for Watermarking.}
Given the black-box nature of diffusion models, direct backpropagation through the perturbation process is infeasible. However, we find an opportunity in contrastive learning, which distinguishes between similar (positive) and dissimilar (negative) data pairs to learn useful representations~\cite{chopra2005learning}. In contrastive learning, augmentations like cropping or color changes are used to train an encoder to identify similar views of the same image while distinguishing those from different images. Interestingly, these augmentations share similarities with the perturbations included in the watermark training process, as they both contain a series of image perturbations like rotation, blur, color changes, etc. This similarity allows us to incorporate diffusion-based perturbations into the contrastive learning paradigm without the need for a differentiable perturbation layer. We adapt this idea to train an encoder-decoder pair to differentiate watermarked images (positive pairs) from non-watermarked ones (negative pairs). We define positive pairs as a watermarked image and its perturbed version, while negative pairs are comprised of the original image and its perturbed version, as well as all wrongly shuffled images. The encoder embeds the watermark, while the decoder is tasked with differentiating between these pairs. The encoder-decoder mechanism aligns with watermarking methods in~\cite{zhu2018hidden, tancik2020stegastamp}, yet we simplify the process by avoiding direct backpropagation through the perturbations during training.


\noindent
\textbf{Jigsaw Embedding for Customizability.}
Contrastive learning enables the training of watermarks without explicit backpropagation through complex processes like diffusion. However, the embedded watermarks are essentially binary, capable of generating only two states: ``watermarked'' or ``non-watermarked.'' This binary design does not allow for embedding a customizable key to quickly identify different stakeholders' IP.

\begin{figure}[!h]
\vspace{-1.5em}
    \begin{center}
    \includegraphics[width=1.0\textwidth]{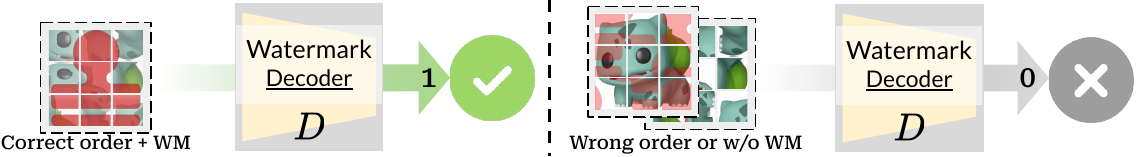}
    \end{center}
    \vspace{-1.7em}
     \caption{Decoder outputs ``1'' only when the watermarked image is in the correct order via the shuffling key. If the order is wrong, despite the presence of the watermark, the decoder outputs ``0''.}
     \vspace{-2.5em}
    \label{fig:Jigsaw}
\end{figure}

To address this limitation, we further introduce the Jigsaw analogy in our watermarking process (Fig.~\ref{fig:Jigsaw}). This method involves shuffling the image before embedding the watermark into the shuffled image. By doing so, the watermark remains intact only when the image is in the specific shuffle order used during embedding. The image is then shuffled back to its original order, effectively hiding the watermark while preserving the image content. During the detection phase, the image must be shuffled using the same order as in the embedding process. The shuffle order thus serves as the watermark key. Only the correct shuffle order can transform the image into the state it was in during watermark embedding, ensuring that the watermark is detectable by the watermark decoder. Incorrect shuffle orders, non-shuffled images, or images without watermarks will cause the decoder to fail in detecting the watermark, as the decoder is trained to classify only intact watermarks as a sign of ``watermarked'' (watermark decoder output is 1).

This information embedding approach is highly capable and efficient. For example, consider an image divided into a 4x4 grid, resulting in 16 jigsaw puzzle blocks. The potential arrangements ($16!$) amount to more than 44 bits of information. Additionally, compared to other watermarking approaches that require fine-tuning~\cite{fernandez2023stable} to embed customized keys, the Jigsaw mechanism can be applied within a few milliseconds, ensuring the efficiency of the watermarking process (see Appendix~\ref{append:more_results} for more details). 

\noindent
\textbf{Loss Functions.}
The efficacy of $\AlgName$ hinges on a composite loss function, expressed as:
$\mathcal{L} = \mathcal{L}_w + \mathcal{L}_v$.
Here, $\mathcal{L}_w$ (the watermark loss) facilitates the decoder's capability to discern between images with correctly shuffled watermarks and those without. Concurrently, $\mathcal{L}_v$ (the visual loss) ensures the watermark's invisibility, preserving the original image's visual quality.

In our model, the decoder outputs two sets of watermark scores: $k_+$ for positive samples and $k_-$ for negative samples. These scores are pivotal in computing $\mathcal{L}_w$, which aims to amplify the distinction between $k+$ and $k_-$, enabling an effective threshold setting for watermark detection during the stage of watermark detection. 

Inspired by contrastive learning, we adapt loss functions from this domain to the problem of watermarking, introducing a novel approach to optimizable watermarking techniques with enhanced robustness. Specifically, we define the watermark loss $\mathcal{L}_w$ using the Temperature Binomial Deviance Loss (TBDL)~\cite{yi2014deep} as follows:

\begin{equation}
\vspace{-.5em}
\mathcal{L}_{w} = \log \left[1+e^{\left(\lambda-k_+\right)/ \tau}\right] + \log \left[1+e^{\left(k_- -\lambda\right)/\tau}\right],
\end{equation}
where $\lambda$ sets the boundary between positive and negative samples, and $\tau$, the temperature, intensifies the model's focus on examples near this boundary, enhancing accuracy.

The visual loss $\mathcal{L}_v$ is adopted to maintain image quality post-watermark embedding. It combines two components:
\begin{equation}
\vspace{-1em}
\mathcal{L}v = \alpha \mathcal{L}_{LPIPS} + \beta \mathcal{L}_{SmoothL1},
\vspace{.2em}
\end{equation}
with the LPIPS loss~\cite{zhang2018unreasonable} assessing perceptual image differences using the VGG model~\cite{simonyan2014very}, and the SmoothL1 loss, which is less outlier-sensitive than MSE, aiding in preserving structural integrity. The coefficients $\alpha$ and $\beta$ balance these components. Further details and ablation studies of these loss functions are detailed in Appendix~\ref{sec:ablation}.

\subsection{Overall Workflow}
In this section, we will introduce the full workflow of $\AlgName$ in detail. For clarity, we use color coding: positive samples (correct watermarked images) are marked in \scalebox{0.9}{\colorbox{mygreen}{\textbf{green}}}, and negative samples (non-watermarked or incorrect watermarked images) in \scalebox{0.9}{\colorbox{mygray}{\textbf{gray}}}. 
We start with the key components of $\AlgName$:

\begin{figure*}[!ht]
    \begin{center}
    \includegraphics[width=1.0\linewidth]{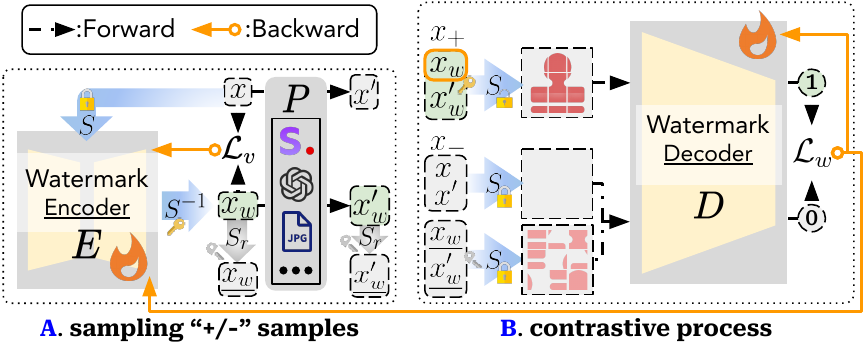}
    \end{center}
    \vspace{-1.em}
     \caption{
     The training process of $\AlgName$ can be seen as two phases, \textcolor{blue}{\textbf{A}}. Phase 1: Sampling positive and negative examples and \textcolor{blue}{\textbf{B}}. Phase 2: Leveraging the difference between the positive and negative samples to train the encoder and decoder via contrastive learning. 
     }
     \vspace{-2em}
    \label{fig:pipeline}
\end{figure*}

\vspace{-.5em}
\begin{packeditemize}
    \item \textbf{Shuffle Rule ($S$)}: segmenting the image into smaller blocks and introducing randomness through shuffling and flipping. The shuffle rule $S$ is defined such that its inverse $S^{-1}$ can accurately reassemble the image into its original configuration.
    
    \item \textbf{Watermark Encoder ($E$)}: The encoder $E$ embeds an imperceptible watermark $w$ into the original image \graybox{$x$} shuffled by $S$. After embedding, with $S^{-1}$ recovering the sematic order, resulting in a watermarked image \greenboxl{$x_w$}. 

    \item \textbf{Perturbations ($P$)}: To mimic perturbation scope encountered in real-world scenarios, we introduce randomized perturbations to image pairs \graybox{$x$} and \greenboxl{$x_w$} during encoder-decoder training, results in perturbed variants, \grayboxt{$x'$} and \greenboxtl{$x'_w$}. $P$ spans a wide range, including diffusion-based image variation, detailed in Appendix~\ref{sec:details}.

    \item \textbf{Watermark Decoder ($D$)}:  The decoder $D$ is designed to interpret images and yield a watermark score $k$, ranging between 0 and 1. This score can be interpreted as the likelihood of watermark presence.
\end{packeditemize}

Training of $\AlgName$ consists of the sampling stage (Fig.~\ref{fig:pipeline} \textcolor{blue}{\textbf{A}}.) and the contrastive learning phase (Fig.~\ref{fig:pipeline} \textcolor{blue}{\textbf{B}}.). The sampling starts with an image, \graybox{$x$}, undergoing a random shuffling operation, $S$, preparing it for watermarking. Post-shuffling, $E$ embeds the watermark into this semantically shuffled image. To revert the image to its original semantic order, $S^{-1}$ is applied, producing the watermarked variant containing the matched Jigsaw shuffling information that can enable the correct order of watermark when $S$ being deployed in the future, \greenboxl{$x_w$}. The fidelity of the watermarking process is quantified by computing the Visual Similarity Loss, $\mathcal{L}_v$, between \graybox{$x$} and \greenboxl{$x_w$}.

Subsequently, both \graybox{$x$} and \greenboxl{$x_w$} undergo a set of random perturbations, $P$, resulting in their perturbed forms \grayboxt{$x'$} and \greenboxtl{$x'_w$}, respectively. Another randomly sampled shuffle operation, $S_r$, further manipulates \greenboxl{$x_w$} and \greenboxtl{$x'_w$} to generate the shuffled states \colorbox[HTML]{d4d4d4}{$\underline{x_w}$} and \grayboxltx{$\underline{x'_w}$}, which represents the watermarked samples that do not shuffle by correct shuffling key. 

In the contrastive process (Fig.~\ref{fig:pipeline} \textcolor{blue}{\textbf{B}}.), samples are classified into positive \greenboxl{$x_+$} (\greenboxl{$x_w$} and \greenboxtl{$x'_w$}) and negative \colorbox[HTML]{d4d4d4}{\textbf{$x-$}} (\grayboxlll{\textbf{$x$}, \textbf{$x'$}, \textbf{$\underline{x_w}$}}, and \grayboxltx{$\underline{x'_w}$}). Samples pass through $S^{-1}$ to reveal the correct watermark order if applicable. Finally, $D$ processes these images, assigning watermark likelihood $k$, and computes the Watermark Loss, $\mathcal{L}_w$.

In $\AlgName$, the combined training loss, composed of $\mathcal{L}_v$ and $\mathcal{L}_w$, undergoes backpropagation to optimize the parameters of $E$ and $D$. Importantly, $\mathcal{L}_w$ serves a dual purpose, guiding the parameter adjustments for both. Specifically, it directs $D$ in distinguishing between positive and negative samples. Meanwhile, for $E$, the gradients are derived from \greenboxl{$x_w$}'s output from $D$ to learn how to inject robust watermarks, as other sample states are involved in non-differentiable transformations. 

A distinctive aspect of $\AlgName$ is its avoidance of direct backpropagation through perturbation processes, boosts its capability to handle complex and non-transparent perturbations, e.g., with closed-source platforms like DALL·E 2. The pseudo-code of the training process is available in Algorithm~\ref{algorithm:pythoncode}, Appendix \ref{sec:jigmarkalg}.

At the deployment stage, $\AlgName$ generates a new, unique shuffle pattern $S'$ for each IP holder who wishes to embed a distinct watermark. The watermarking process begins with an image being shuffled using $S'$, then passed through $E$ for watermark embedding, followed by unshuffling with $S'^{-1}$. $D$ then assesses the image to calculate the watermark likelihood, $k$. If the inverse shuffle pattern used during deployment differs from the one used at the encoding stage, the decoder $D$ will yield a low value of $k$, indicating a mismatch in the watermarking process (Fig.~\ref{fig:Jigsaw}).

\section{Evaluation}
\label{sec:mainevaluation}
We evaluate $\AlgName$ against the diverse real-world challenges detailed in our threat models (Section~\ref{sec:threatmodels}). Our evaluation focuses on the following aspects:
\begin{packeditemize}
\item \textbf{Robustness:} (Section~\ref{sec:case1exp}) We examine how $\AlgName$ compares to other baseline watermarking methods under \textbf{Types 1, 2, and 3} perturbations.
\item \textbf{Visual Stealthiness:} (Section~\ref{sec:viscompare}) We evaluate the visual impact of $\AlgName$ against other watermarking methods on an image.
\item \textbf{Fasle Positive Rate:} (Appendix~\ref{append: personalkey}) We analyze the similar watermark key misclassification rate of $\AlgName$ and other baselines.
\end{packeditemize}
Further evaluations and the ablation study are deferred to Appendix~\ref{append:more_results}.

\subsection{A Human-Aligned Image Variation ($\SimName$) Metric}
\label{sec:hav}

Evaluating the performance of watermarking methods against diffusion model-based image modifications requires a suitable metric that aligns with human perception. Existing image similarity metrics, such as MSE, SSIM~\cite{wang2004image}, PHash~\cite{zauner2010implementation}, LPIPS~\cite{zhang2018unreasonable}, and CLIP~\cite{radford2021learning}, have limited effectiveness in capturing and quantify these complex image variations, as we presented in Table~\ref{tab:viseval}.

\begin{table}[th!]
\centering
\vspace{-1em}
\resizebox{0.70\textwidth}{!}{ 
\begin{tabular}{l|c|ccccc|c} 
& Human &  CLIP & LPIPS & PHash & SSIM & MSE & \textbf{HAV(Ours)}\\ \hline
\textbf{Spearman($\downarrow$)} & 2.56 & 6.81 & 7.06 & 7.21 & 7.11 & 7.92 & \textbf{2.89} \\ 
\end{tabular}
}
\vspace{0.2em}
\caption{Spearman distance of rankings based on different image similarity metrics vs. rankings based on human annotations. Human is the average distance to other annotator assessments in leave-one-out evaluation (see Appendix~\ref{sec:modified_dataset}).}
\vspace{-2.5em}
\label{tab:viseval}
\end{table}

To address this limitation, we introduce the Human-Aligned Variation (HAV) score, a metric developed directly from human annotations. We collected human rankings for 2,200 image groups, each containing an original image and its modified versions generated by different diffusion models (detailed in Appendix~\ref{sec:modified_dataset}). We then trained a Siamese Network~\cite{bromley1993signature} on this data (detailed in Appendix~\ref{sec:humansim}) to predict HAV scores ranging from 0.0 (low modification) to 1.0 (high modification).

The HAV score achieves a Spearman Distance of 2.89 to human ranking vectors, closely aligning with the average discrepancy in rankings between different annotators (2.56). This indicates that the HAV score effectively captures the degree of similarity typically observed between human assessments.

\begin{figure}[htbp]
\vspace{-1.0em}
    \begin{center}
    \includegraphics[width=1.00\textwidth]{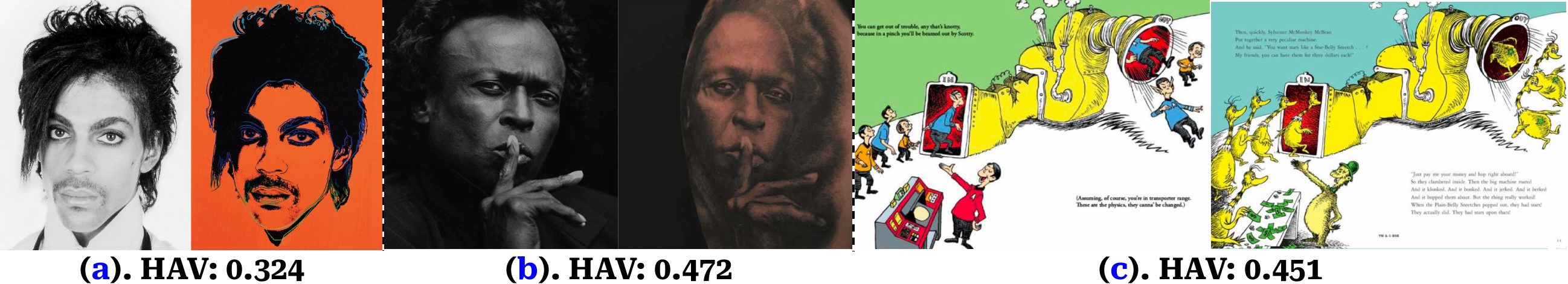}
    \end{center}
    \vspace{-1.7em}
     \caption{Analysis of artwork cases deemed as plagiarized in court.
     (\textcolor{blue}{\textbf{a}}) Andy Warhol Foundation for Visual Arts, Inc. v. Goldsmith~\cite{warhol_2023},
     (\textcolor{blue}{\textbf{b}}) Sedlik v. Von Drachenberg~\cite{sedlik_von_drachenberg_2022},
     (\textcolor{blue}{\textbf{c}}) Dr. Seuss Enterprises, L.P. v. ComicMix LLC~\cite{seuss_vs_comicmix}.
     }
    \label{fig:lawcase}
    \vspace{-2em}
\end{figure}

To validate the practical utility of the HAV score, we applied it to analyze prominent copyright infringement cases involving significant image transformations (Fig.~\ref{fig:lawcase}). We found that a threshold of HAV $\leq$ 0.5 successfully captured all cases where the court ruled that a derivative work was not transformative enough to qualify for a fair use defense (more discussion in Appendix~\ref{app:law}). This threshold serves as a benchmark in our evaluations, particularly for assessing the robustness of watermarking methods against noticeable image alterations.

In our following evaluation, we use the $\SimName$ score to quantify the strength of the image editing. We generate modified images using various diffusion models and filter them based on their $\SimName$ scores, keeping only those within the range of 0.3 to 0.5. This ensures that the evaluated perturbations represent noticeable modifications while remaining below the legally-informed 0.5 threshold. During the experiments, we use these $\SimName$-filtered images to measure the robustness of $\AlgName$ and other baseline watermark methods. By incorporating the $\SimName$ score as an evaluation metric, we align our assessment of watermarking methods with human perception and real-world legal considerations, demonstrating the effectiveness of $\AlgName$ in preserving watermark detectability under complex, diffusion model-based image modifications.

\subsection{Settings}
\noindent
\textbf{Baselines Watermarks.} We benchmark $\AlgName$ against traditional frequency-based watermark DctDwtSVD~\cite{4554423}, deep learning-based HiDDeN~\cite{zhu2018hidden}, and diffusion model-integrated watermarks Stable Signature~\cite{fernandez2023stable} and Tree-Ring~\cite{wen2023tree}. For a fair comparison, we fixed the secret bit size to 44 bits across all watermarking methods. In the case of Tree-Ring, we use the $\text{Tree-Ring}_{\textbf{Rings}}$ variant. For $\AlgName$, we employ 16 blocks, which translates to approximately 44 bits of information, to maintain consistency with the other methods.

\noindent\textbf{Perturbation Types.} 
Consistent with the threat models (Section~\ref{sec:threatmodels}, we test against \textbf{Type 1, 2, and 3} perturbations, encompassing common distortions (JPEG compression, Gaussian Blur), diffusion model variations (SDEdit~\cite{meng2021sdedit}, and unseen ones, DALL·E 2~\cite{ramesh2022hierarchical}, InstructPix2Pix~\cite{brooks2023instructpix2pix}, Zero 1-to-3~\cite{liu2023zero}, InPaint~\cite{lugmayr2022repaint}). In evaluating \textbf{Type 2} perturbations, we apply $ 0.3 \leq \SimName \leq 0.5$. This range is selected to ensure noticeable changes to the image while maintaining the threshold of $\leq 0.5$ that as discussed in Section~\ref{sec:hav}. Additionally, \textbf{Type 3} watermark removal attacks, including RG~\cite{zhao2023invisible}, WEvade-B-Q~\cite{jiang2023evading}, AdvH~\cite{saberi2023robustness}, and AC~\cite{lukas2023leveraging}, are implemented in a black-box setting without direct decoder access, while the PGD ~\cite{madry2017towards} is evaluated as a white-box setting. Detailed hyper-parameters and settings are available in Appendix~\ref{sec:details}.
We also include the other generative model as image perturbation such as GAN~\cite{isola2017image} on Appendix~\ref{append:more_results}.

\begin{table*}[t!]
\centering
\resizebox{1.0\textwidth}{!}{
\begin{tabular}{c|cc|cc|cc|cc|cc|cc}
\hline
& \multicolumn{2}{c|}{\textbf{JPEG}} & \multicolumn{2}{c|}{\textbf{Gaussian Noise}} & \multicolumn{2}{c|}{\textbf{Gaussian Blur}} & \multicolumn{2}{c|}{\textbf{Rotation}} & \multicolumn{2}{c|}{\textbf{Contrast \& Brightness}} & \multicolumn{2}{c}{\textbf{Average}}\\ \hline
& \multicolumn{1}{c}{\textbf{\begin{tabular}[c]{@{}c@{}}TPR ($\uparrow$) @\\ 1\%FPR\end{tabular}}} & \textbf{AUC ($\uparrow$)} & \multicolumn{1}{c}{\textbf{\begin{tabular}[c]{@{}c@{}}TPR ($\uparrow$) @\\ 1\%FPR\end{tabular}}} & \textbf{AUC ($\uparrow$)} & \multicolumn{1}{c}{\textbf{\begin{tabular}[c]{@{}c@{}}TPR ($\uparrow$) @\\ 1\%FPR\end{tabular}}} & \textbf{AUC ($\uparrow$)} & \multicolumn{1}{c}{\textbf{\begin{tabular}[c]{@{}c@{}}TPR ($\uparrow$) @\\ 1\%FPR\end{tabular}}} & \textbf{AUC ($\uparrow$)} & \multicolumn{1}{c}{\textbf{\begin{tabular}[c]{@{}c@{}}TPR ($\uparrow$) @\\ 1\%FPR\end{tabular}}} & \textbf{AUC ($\uparrow$)} & \multicolumn{1}{c}{\textbf{\begin{tabular}[c]{@{}c@{}}TPR ($\uparrow$) @\\ 1\%FPR\end{tabular}}} & \textbf{AUC ($\uparrow$)} \\ \hline
\textbf{\begin{tabular}[c]{@{}c@{}}DwtDctSVD\\ \cite{4554423}\end{tabular}} & \multicolumn{1}{c}{\large 0.962} & \large 0.989 & \multicolumn{1}{c}{\large 0.280} & \large 0.841 & \multicolumn{1}{c}{\large 0.806} & \large 0.963 & \multicolumn{1}{c}{\large 0.542} & \large 0.913 & \multicolumn{1}{c}{\large 0.243} & \large 0.684 & \multicolumn{1}{c}{\large 0.567} & \large 0.878 \\ \hline
\textbf{\begin{tabular}[c]{@{}c@{}}HiDDeN\\ \cite{zhu2018hidden}\end{tabular}} & \multicolumn{1}{c}{\large 0.972} & \large 0.997 & \multicolumn{1}{c}{\large 0.482} & \large 0.903 & \multicolumn{1}{c}{\large 0.774} & \large 0.958 & \multicolumn{1}{c}{\large 0.937} & \large 0.998 & \multicolumn{1}{c}{\large 0.802} & \large 0.964 & \multicolumn{1}{c}{\large 0.793} & \large 0.964 \\ \hline
\textbf{\begin{tabular}[c]{@{}c@{}}Stable Sig*\\ \cite{fernandez2023stable}\end{tabular}} & \multicolumn{1}{c}{\large 0.770} & \large 0.955 & \multicolumn{1}{c}{\large 0.330} & \large 0.858 & \multicolumn{1}{c}{\large 0.740} & \large 0.949 & \multicolumn{1}{c}{\large 0.883} & \large 0.984 & \multicolumn{1}{c}{\large 0.582} & \large 0.911 & \multicolumn{1}{c}{\large 0.661} & \large 0.931 \\ \hline
\textbf{\begin{tabular}[c]{@{}c@{}}Tree-Ring*\\\cite{wen2023tree} \end{tabular}} & \multicolumn{1}{c}{\large 0.986} & \large \textbf{0.999} & \multicolumn{1}{c}{\large 0.438} & \large 0.887 & \multicolumn{1}{c}{\large 0.901} & \large 0.988 & \multicolumn{1}{c}{\large \textbf{0.988}} & \large \textbf{0.999} & \multicolumn{1}{c}{\large 0.658} & \large 0.944 & \multicolumn{1}{c}{\large 0.794} & \large 0.963 \\ \hline
\textbf{\begin{tabular}[c]{@{}c@{}}$\AlgName$\\ (Ours)\end{tabular}} & \multicolumn{1}{c}{\large \textbf{0.992}} & \large 0.997 & \multicolumn{1}{c}{\large \textbf{0.932}} & \large \textbf{0.992} & \multicolumn{1}{c}{\large \textbf{0.994}} & \large \textbf{0.999} & \multicolumn{1}{c}{\large 0.982} & \large 0.998 & \multicolumn{1}{c}{\large \textbf{0.879}} & \large \textbf{0.983} & \multicolumn{1}{c}{\large \textbf{0.956}} & \large \textbf{0.994} \\ \hline
\end{tabular}
}
\caption{Comparison of watermark detection under traditional image perturbations. *For Stable Sig and Tree-Ring, results are derived from their synthetic data generated using textual guidance from our evaluation dataset (image-text pairs).}
\label{tab:stage1res}
\vspace{-1.5em}
\end{table*}

\begin{table*}[t!]
\centering
\resizebox{1.0\textwidth}{!}{
\begin{tabular}{c|cc|cc|cc|cc|cc|cc}
\hline
                          & \multicolumn{2}{c|}{\textbf{SDEdit\cite{meng2021sdedit}}}                                                                & \multicolumn{2}{c|}{\textbf{InstructPix2Pix\cite{brooks2023instructpix2pix}}}                                                         & \multicolumn{2}{c|}{\textbf{Zero 1-to-3\cite{liu2023zero}}}                                                          & \multicolumn{2}{c|}{\textbf{InPaint\cite{lugmayr2022repaint}}}                                                               & \multicolumn{2}{c|}{\textbf{DALL·E 2\cite{ramesh2022hierarchical}}} & \multicolumn{2}{c}{\textbf{Average}}                                                              \\ \hline
                          & \multicolumn{1}{c}{\textbf{\begin{tabular}[c]{@{}c@{}}TPR ($\uparrow$) @\\ 1\%FPR\end{tabular}}} & \textbf{AUC ($\uparrow$)}   & \multicolumn{1}{c}{\textbf{\begin{tabular}[c]{@{}c@{}}TPR ($\uparrow$) @\\ 1\%FPR\end{tabular}}} & \textbf{AUC ($\uparrow$)}   & \multicolumn{1}{c}{\textbf{\begin{tabular}[c]{@{}c@{}}TPR ($\uparrow$) @\\ 1\%FPR\end{tabular}}} & \textbf{AUC ($\uparrow$)}  & \multicolumn{1}{c}{\textbf{\begin{tabular}[c]{@{}c@{}}TPR ($\uparrow$) @\\ 1\%FPR\end{tabular}}} & \textbf{AUC ($\uparrow$)}   & \multicolumn{1}{c}{\textbf{\begin{tabular}[c]{@{}c@{}}TPR ($\uparrow$) @\\ 1\%FPR\end{tabular}}} & \textbf{AUC ($\uparrow$)} & \multicolumn{1}{c}{\textbf{\begin{tabular}[c]{@{}c@{}}TPR ($\uparrow$) @\\ 1\%FPR\end{tabular}}} & \textbf{AUC ($\uparrow$)}  \\ \hline
\textbf{\begin{tabular}[c]{@{}c@{}}DwtDctSVD\\ \cite{4554423}\end{tabular}}        & \multicolumn{1}{c}{\large 0.020}                                                         & \large 0.537          & \multicolumn{1}{c}{\large 0.030}                                                          & \large 0.500          & \multicolumn{1}{c}{\large 0.010}                                                         & \large 0.510         & \multicolumn{1}{c}{\large 0.010}                                                         & \large 0.586          & \multicolumn{1}{c}{\large 0.010}                                                         & \large 0.500          & \multicolumn{1}{c}{\large 0.014}                                                & \large 0.527\\ \hline
\textbf{\begin{tabular}[c]{@{}c@{}}HiDDeN\\ \cite{zhu2018hidden}\end{tabular}}           & \multicolumn{1}{c}{\large 0.020}                                                         & \large 0.521          & \multicolumn{1}{c}{\large 0.042}                                                         & \large 0.619          & \multicolumn{1}{c}{\large 0.010}                                                        & \large 0.592         & \multicolumn{1}{c}{\large 0.030}                                                         & \large 0.689          & \multicolumn{1}{c}{\large 0.000}                                                         & \large 0.589          & \multicolumn{1}{c}{\large 0.021}                                               & \large 0.602\\ \hline
\textbf{\begin{tabular}[c]{@{}c@{}}Stable Sig*\\ \cite{fernandez2023stable}\end{tabular}} & \multicolumn{1}{c}{\large 0.010}                                                         & \large 0.510          & \multicolumn{1}{c}{\large 0.010}                                                          & \large 0.589          & \large 0.000 & \large 0.500                                                                           & \multicolumn{1}{c}{\large 0.030}                                                         & \large 0.629          & \multicolumn{1}{c}{\large 0.010}                                                         & \large 0.561          & \multicolumn{1}{c}{\large 0.012}                                                & \large 0.558\\ \hline
\textbf{\begin{tabular}[c]{@{}c@{}}Tree-Ring*\\\cite{wen2023tree} \end{tabular}}        & \multicolumn{1}{c}{\large 0.143}                                                         & \large 0.880          & \multicolumn{1}{c}{\large 0.133}                                                         & \large 0.826          & \large 0.107 &\large 0.794                                                                            & \multicolumn{1}{c}{\large 0.218}                                                         & \large 0.895          & \multicolumn{1}{c}{\large 0.231}                                                         & \large 0.856          & \multicolumn{1}{c}{\large 0.166}                                                & \large 0.850\\ \hline
\textbf{\begin{tabular}[c]{@{}c@{}}$\AlgName$\\ (Ours)\end{tabular}}             & \multicolumn{1}{c}{\textbf{\large 0.945}}                                                & \textbf{\large 0.989} & \multicolumn{1}{c}{\textbf{\large 0.953}}                                                & \textbf{\large 0.992} & \multicolumn{1}{c}{\textbf{\large 0.876}}                                                 & \textbf{\large 0.981} & \multicolumn{1}{c}{\textbf{\large 0.873}}                                                & \textbf{\large 0.975} & \multicolumn{1}{c}{\textbf{\large 0.721}}                                                & \textbf{\large 0.934} & \multicolumn{1}{c}{\textbf{\large 0.874}}                                                & \textbf{\large 0.974}\\ \hline
\end{tabular}

}
\caption{Comparison of watermark detection under various diffusion-based image editing techniques. *For Stable Sig and Tree-Ring, results are derived from their synthetic data generated using textual guidance from our evaluation datasets (image-text pairs).}
\vspace{-3.0em}
\label{tab:stage2resmin0.3}
\end{table*}

\noindent 
\textbf{Evaluation Metrics.} Our analysis employs three key metrics:
\textit{Area Under the Curve (AUC)}: Assesses the watermark system's discernment between watermarked and non-watermarked images, with higher AUC reflecting greater effectiveness and perturbation robustness.
\textit{True Positive Rate (TPR) at 1\% False Positive Rate (FPR)}: Measures the accuracy of watermark detection, maintaining a balance between identifying true positives and minimizing false positives.
\textit{Bit Correct Ratio (BCR)}: For multi-bit watermarks, BCR evaluates the accuracy of watermark key recovery, ranging from 0 (complete failure) to 1 (perfect recovery).
\textit{Attack Success Rate (ASR)}: For watermark attacks, ASR quantifies the effectiveness of watermark removal attacks. It reflects the proportion of watermarked examples that are successfully altered to be classified as non-watermarked.

\noindent
\textbf{Dataset for Evaluation.} Recognizing the necessity of image-text pairs (using the text to create image-related but random instructions) for evaluating diffusion-based perturbations, we opt not to use common datasets like LAION-5B~\cite{schuhmann2022laion} or InstructPix2Pix~\cite{brooks2023instructpix2pix} (which is based on LAION-5B), to avoid data leakage (we still include the results over it in Appendix \ref{append:diffdataset} just for reference), as these datasets have been extensively used in model training. Instead, we use the ImageNet-1k dataset~\cite{5206848}, appending it with newly created textual descriptions. This approach, detailed in Appendix~\ref{sec:modified_dataset}, involves selecting 2,000 image-text pairs from the ImageNet-1k validation set, labeled with LLaVA~\cite{liu2023visual}. For assessments involving diffusion-integrated watermarks~\cite{fernandez2023stable,wen2023tree}, evaluations are conducted with synthetic data generated using these captions (of the 2000 images).

\subsection{Robustness Evaluation}
\textbf{Type 1 - Conventional Perturbations.}
\label{sec:case1exp}
Table~\ref{tab:stage1res} reveals varying performances of watermarking methods against traditional distortions. DwtDctSVD~\cite{4554423} exhibits limited effectiveness, particularly under Gaussian noise and contrast adjustments, due to its reliance on hand-crafted triggers and vulnerability in the high-frequency domain. HiDDeN~\cite{zhu2018hidden} and Stable Signature~\cite{fernandez2023stable} show comparable results; however, Stable Signature's performance is slightly diminished by its focus on fine-tuning the latent decoder alone.
Tree-Ring~\cite{wen2023tree} achieves high AUC across most tests but struggles with Gaussian noise, a direct consequence of its watermark embedding in the diffusion model's latent space, which is sensitive to such noise.
$\AlgName$, in contrast, consistently outperforms others in this category, demonstrating its robustness against a variety of traditional distortions.

\label{sec:viscompare}
\begin{figure*}[!t]
    \centering
    \includegraphics[width=1.0\textwidth]{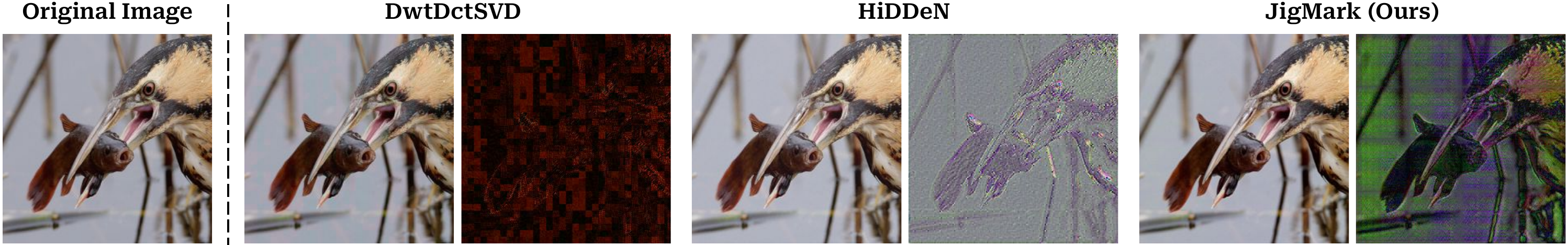}
    \vspace{-1.5em}
    \caption{Visual comparison of watermarking techniques: watermarked images on the left and magnified deficits (×10) on the right. $\AlgName$ distinctively embeds low-frequency noise in less noticeable areas like boundaries and textures
}
    \vspace{-2em}
    \label{fig:visual_comparison}
\end{figure*}

\noindent
\textbf{Type 2 - Diffusion Perturbations.}
\label{sec:case2exp}
Within the evaluated range of moderate level of diffusion-based image variations (HAV score ranging from 0.3 to 0.5), Table~\ref{tab:stage2resmin0.3} shows that DwtDctSVD~\cite{4554423}, HiDDeN~\cite{zhu2018hidden}, and Stable Signature~\cite{fernandez2023stable} struggle with watermark detection under diffusion perturbations, with AUCs near 0.5, implying performance akin to random guessing. DwtDctSVD's low performance is attributed to the distortion of its high-frequency space watermark under diffusion processes. Both HiDDeN and Stable Signature, designed for linear approximations of perturbations, falter against the complex modifications introduced by modern generative models.
Tree-Ring~\cite{wen2023tree}, although it demonstrates better robustness than the other baselines, the TPR at 1\% FPR results are still of a low level, indicating their limitation in facing the evaluated moderate level of diffusion perturbations.
Conversely, $\AlgName$ maintains high AUCs across all evaluated diffusion perturbations.

Significantly, $\AlgName$ was trained using only SDEdit-processed samples with a variety of arbitrary prompts (see Appendix \ref{sec:details}), yet it exhibited exceptional robustness against perturbations from diffusion models it hadn't encountered before. This adaptability may be attributed to the commonality of the diffusion process used in these models. As a side note, we can further finetune the trained encoder-decoder pairs of $\AlgName$ on DALL·E 2. The finetuned $\AlgName$ achieves an AUC of 0.98 and a TPR of 0.824. This process is further detailed in Appendix~\ref{append:more_results}, highlighting $\AlgName$'s capability to adapt and respond efficiently to new and unseen perturbations.

\vspace{-1.5em}
\def\arraystretch{1.2}
\begin{table}[h]
\centering
\resizebox{0.55\columnwidth}{!}{
\begin{tabular}{c|c|c|c|c|c|c}
\hline
& \textbf{ RG~\cite{zhao2023invisible}} & \textbf{ WEv\cite{jiang2023evading}} & \textbf{ AdvH~\cite{saberi2023robustness}} & \textbf{ AC~\cite{lukas2023leveraging}} & \textbf{ PGD$^{\Diamond}$~\cite{madry2017towards}} & \textbf{Average} \\ \hline
\textbf{ \begin{tabular}[c]{@{}c@{}}DwtDctSVD\\ \cite{4554423}\end{tabular}} & \large 84.21 & \large 27.53 & \large 73.31 & \large 95.72 & \large 97.21 & \large 76.20 \\ \hline
\textbf{ \begin{tabular}[c]{@{}c@{}}HiDDeN\\ \cite{zhu2018hidden}\end{tabular}} & \large 78.52 & \large 44.32 & \large 67.63 & \large 98.52 & \large 98.57 & \large 77.51 \\ \hline
\textbf{ \begin{tabular}[c]{@{}c@{}}Stable Sig*\\ \cite{fernandez2023stable}\end{tabular}} & \large 81.43 & \large 47.31 & \large 69.81 & \large 97.69 & \large 96.43 & \large 78.53 \\ \hline
\textbf{ \begin{tabular}[c]{@{}c@{}}Tree-Ring*\\\cite{wen2023tree} \end{tabular}} & \large \textbf{2.77} & \large \textbf{2.21} & \large 88.21 & \large 64.53 & \large 99.28 & \large 51.40 \\ \hline
\textbf{ \begin{tabular}[c]{@{}c@{}}$\AlgName$\\ (Ours)\end{tabular}} & \large 3.10 & \large 5.21 & \large \textbf{35.42} & \large \textbf{37.61} & \large \textbf{91.27} & \large \textbf{34.52} \\ \hline
\end{tabular}}
\caption{ASR comparison via watermark removal attacks. The $\Diamond$ indicates that the attack method can have white-box access to the watermark model. *For Stable Sig and Tree-Ring, results are derived from their synthetic data generated using textual guidance from our evaluation dataset (image-text pairs).}
\vspace{-2.5em}
\label{tab:stage3}
\end{table}

\noindent\textbf{Type 3 - Watermark Removal Attacks.}
\label{sec:case3exp} 
Table~\ref{tab:stage3} shows RG~\cite{zhao2023invisible}, using a diffusion model-based method, effectively disrupts both frequency and deep learning-based watermarks, echoing the vulnerabilities observed in Type 2 perturbations. WEvade's effectiveness, even with limited queries, is notable against HiDDeN and Stable Sig due to its strategic manipulation of crucial image features for watermark detection. Conversely, Tree-Ring's simplicity in latent space watermark design makes it susceptible to AdvH~\cite{saberi2023robustness}, an adversarial attack. The white-box attack PGD demonstrates high ASRs against all methods, including $\AlgName$, albeit to a slightly lesser extent. However, PGD's requirement for decoder access limits its real-world applicability. The overarching insight is that exposure of decoder details to adversarial attacks poses a significant risk, emphasizing the need for secrecy in decoder design and operation, even for robust watermarking methods like Tree-Ring and ours.

\subsection{Stealthiness Evaluation}

Fig.~\ref{fig:visual_comparison} presents a visual comparison of watermarked samples using $\AlgName$ against other baseline watermarking methods. $\AlgName$ stands out for its enhanced robustness in watermark detection while maintaining stealthiness. This ensures that the watermarks are imperceptible to the human eye, leaving no conspicuous traces of the trigger mechanism. Additionally, for a more in-depth understanding, we provide a quantitative evaluation of the watermarking performance using various image similarity metrics (PSNR, SSIM~\cite{wang2004image}, LPIPS~\cite{zhang2018unreasonable}). These detailed analyses are available in Appendix~\ref{append:more_results}.

\section{Discussion \& Conclusion}

\noindent
\textbf{Limitation and Training Overhead.} While $\AlgName$ demonstrates robustness against diverse perturbations, including those generated by generative models, it does come with significant training demands. The requirement for approximately 1000 hours on an A100 GPU is notably higher than traditional watermarking methods. However, this investment in training is a one-time effort, with subsequent fine-tuning being more resource-efficient and offering adaptability to new perturbations (DALL·E 2\cite{ramesh2022hierarchical} example in Appendix~\ref{append:more_results}).

\noindent
\textbf{Impact on Data Integrity and Copyright Protection.} $\AlgName$ marks a significant step forward in reliable watermarking, crucial for maintaining data integrity amidst the proliferation of synthetic content. It offers content creators and rights-holders a practical tool to identify and protect against unauthorized derivative works, aligning detection mechanisms with human perceptions of image similarity (further elaborated in Appendix~\ref{app:law}). This alignment is particularly pertinent as we navigate the challenges posed by advanced AI technologies in the realm of IP rights.

\noindent
\textbf{Contribution and Advancements.} Our research is the first to demonstrate the vulnerabilities of existing watermarking techniques against diffusion model-based image editing. As a solution, we propose $\AlgName$, a novel optimizable watermark framework that can be trained without requiring perturbation gradients. $\AlgName$ achieves robustness against diffusion model-based image modifications by incorporating the diffusion model editing process into the training. 

\clearpage  

%
%
\bibliographystyle{splncs04}
\bibliography{main}

\clearpage
\appendix
{\centering
\textbf{\thetitle}\\
\vspace{0.5em}Supplementary Material \\}

\section{Broaden Impact \& Scope}
\vspace{-.5em}
\textbf{Diffusion-robust Watermark's Impact.} 
\addcontentsline{toc}{subsection}{Diffusion-robust Watermark's Impact.}
Watermarking has long been essential in protecting IP rights, especially for content in image formats. It acts as a safeguard against unauthorized use, derivation, and exploitation of such materials \cite{uscode2023}. However, with the advent of advanced generative models grounded in diffusion models, reliable watermarks can help in cases where they serve critical responsibilities. These include mitigating financial and reputational damages for creators \cite{Francis_2023}, curbing the proliferation of misinformation \cite{Cheetham_2023}, protecting personal privacy \cite{salman2023raising}, and maintaining the authenticity and reliability of content within the AI data ecosystem \cite{alemohammad2023selfconsuming}. In this evolving digital landscape, the role of watermarking extends beyond traditional IP protection, can be serve as a vital tool in ensuring the ethical use and dissemination of digital content. To motivate future research, we open-source our code.

\noindent\textbf{Unseen and Zero-day Perturbations.} 
\addcontentsline{toc}{subsection}{Unseen and Zero-day Perturbations.}
$\AlgName$ marks an advancement in tracking derivative content. Nevertheless, it is not impervious to new kinds of perturbations. We highlight a key strength of our approach--adaptability; $\AlgName$ can incorporate emerging forms of perturbations into subsequent tuning phases, thereby fortifying the watermark's resilience (Appendix \ref{append:dalle2} exemplifies how to adapt to improve robustness to DALL$\cdot$E 2 image variation).

\noindent\textbf{ControlNet is out of Scope.} 
\addcontentsline{toc}{subsection}{ControlNet is out of Scope.}
Some recent image conditional work, such as ControlNet~\cite{zhang2023adding}, utilizes various image-related information, like edge details and human poses, as conditions to control the image generation process. This method significantly differs from our scope as it primarily pertains to the text-to-image generation domain. ControlNet introduces a control mechanism that manipulates the generative process based on textual inputs, guiding the production of images to match specific desired attributes. In contrast, our study concentrates on watermarking techniques within image generation and manipulation, particularly in image-to-image diffusion models such as SDEdite~\cite{meng2021sdedit} and InstructPix2pix~\cite{brooks2023instructpix2pix}. Given that ControlNet operates on a different axis - influencing the creation of images from text, rather than altering existing images - it falls outside our threat model and the scope of this work. 

\vspace{-0.5em}
\section{Law \& Policy Discussion}
\vspace{-.5em}
\label{app:law}

In copyright litigation, rights-holders argue some notion of similarity at two potential stages (among others that we will not discuss here). First, rights holders use the ``substantial similarity'' test to determine whether the original work was used for some derivative work. This test requires determining whether the defendant had access to the original work to create a derivative and whether the derivative is so similar as to be infringing~\cite{asay2022empirical,asay2020independent}. 
Second, defendants will argue a \textit{fair use} defense, typically saying that the work itself, or the use of the work, is so transformative as to be permissible under the law.
Both analyses are subjective and human-centric---notoriously so---and it is difficult to come up with a precise metric of similarity that would yield consistent accuracy for predicting court outcomes. 
Nonetheless, rights-holders must crawl the web and identify cases where their work has been used in an impermissible way. In many cases rights-holders are entitled to compensation for derivative works beyond simple exact matching. As \cite{henderson2023foundation} discuss, transformations that would not be caught be exact or fuzzy matching may nonetheless be infringing works.
Instead, \cite{henderson2023foundation} propose that future research should invest in human-centric measures of similarity.
Our work provides a step forward, developing human-centric method for identifying transformations that might not be fair use (i.e., leveragting the proposed $\SimName$ score, Appendix \ref{sec:humansim}). Since there is no exact threshold, our methods can be calibrated so that rights-holders can identify the scope of transformations that they wish to detect.
To be clear, this work is not foolproof. Extreme transformations (for example resetting the image to random initialization and then running diffusion) could still remove watermarks.
This is why providing a calibration threshold for rights-holders, aligned with human expectations is so important. 
Rights-holders will \textit{want} to capture some degree of transformation from the original work, and ensure that a watermark withstands this set of transformations, but they will not necessarily want to capture extreme transformations that will easily be defensible in court.
As rights-holders increasingly worry that their work is used by AI without their permission in ways that are not defensible under fair use doctrine, our work can identify pieces of content that align with potential public perceptions of similarity that would be a centerpiece of subsequent litigation.

\subsection{Cases Examined}
We examine several cases in the main text where our similarity metric is applied. Andy Warhol Foundation for Visual Arts, Inc. v. Goldsmith~\cite{warhol_2023}, Sedlik v. Von Drachenberg~\cite{sedlik_von_drachenberg_2022}, and Dr. Seuss Enterprises, L.P. v. ComicMix LLC~\cite{seuss_vs_comicmix}. In all of these cases courts ruled that the transformation in question was \textit{not} fair use. In all cases, a number of other factors were considered and fair use is not always assessed by the level of transformation. Nonetheless, if the downstream derivative works had been sufficiently transformative they would have been more likely to succeed in their fair use defense.

\section{Human Aligned Variation Scores Details}
\label{sec:humansim}
In this Section, we explore advanced image similarity evaluation techniques tailored for images modified by generative AI tools. We detail our approach starting from data collection involving human annotators, to the training of a specialized neural network model assigning the $\SimName$ scores, and finally, an in-depth evaluation comparing our methodology with established benchmarks in the field.
\begin{figure*}[h]
    \centering
    \includegraphics[width=0.98\linewidth]{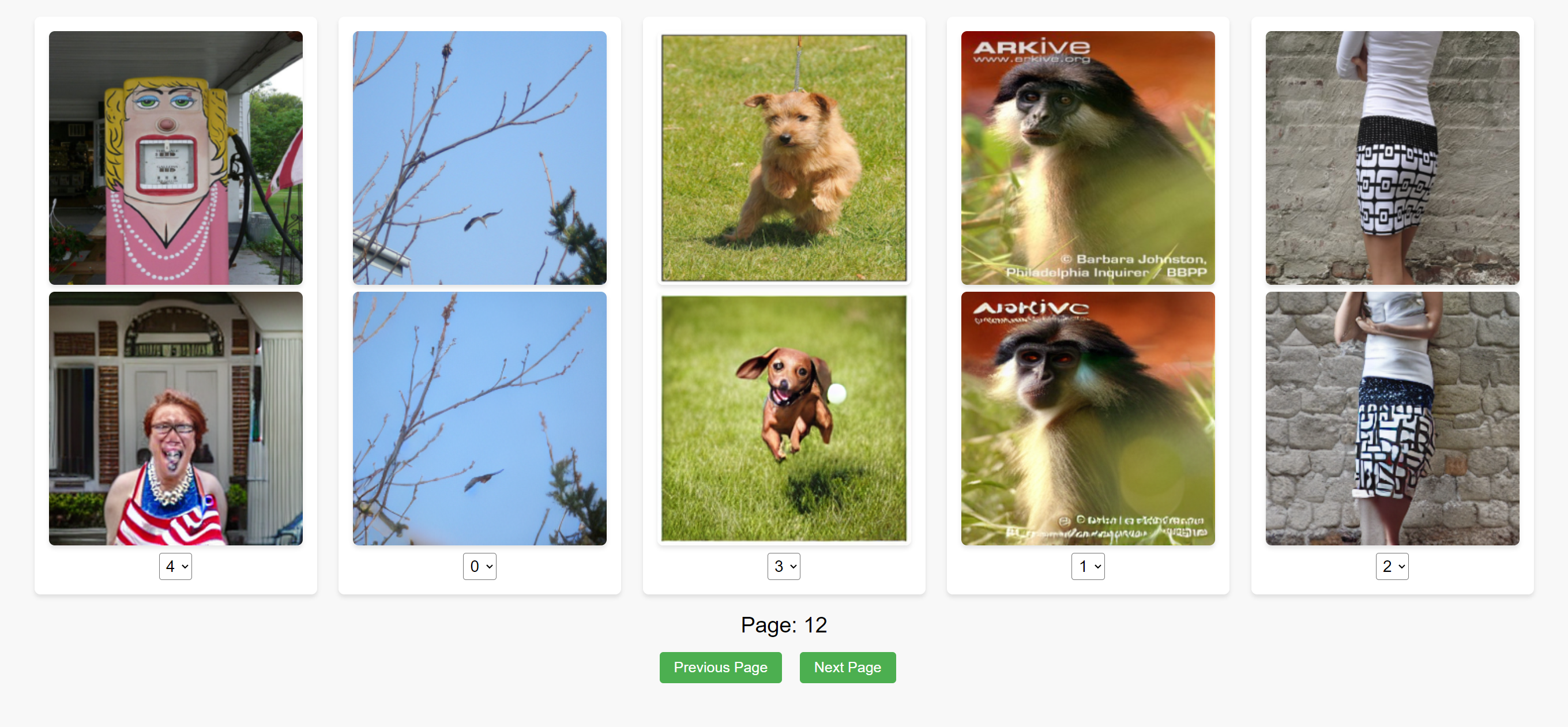}
    \vspace{-.5em}
    \caption{The user interface designed for image variation ranking. In each set, the upper image serves as the reference, while the lower images represent derivatives. Human annotators are tasked to assign rankings, where ``0'' denotes the highest similarity and ``4'' the least.}
    \vspace{-1em}
    \label{fig:ranking}
\end{figure*}

\noindent\textbf{Data Collection for $\SimName$.}
\addcontentsline{toc}{subsection}{Data Collection for $\SimName$.}
Traditional image similarity metrics, such as MSE, SSIM~\cite{wang2004image}, and LPIPS~\cite{zhang2018unreasonable}, predominantly quantify semantically irrelevant changes or manually imperceptible changes. Notably, they often fail to capture image similarity when the visual content undergoes intricate and profound modifications per our evaluation in Section \ref{sec:hav}. We uncover the limitations of existing image similarity metrics in depicting information derivatives that align with humans. Some recent works have proposed training models directly on human-annotated data to measure the synthetic data image quality~\cite{wu2023better, kirstain2023pick, wu2023human}. 
Following these methods, to establish an accurate image variation metric, our initial step involved data collection from annotators. An intuitive approach is to present a pair of images – an original and its modified counterpart – and then solicit annotations on their similarity as a similarity score. However, a challenge surfaces when we recognize that humans might struggle to maintain a consistent standard across a multitude of images.
To circumvent this potential inconsistency, we reframe the scoring task as a ranking problem. As depicted in Fig.~\ref{fig:ranking}, our tailored user interface presents five pairs of images: the original and its altered version. Human annotators are then tasked with ranking these pairs based on perceived similarity; a rank of 0 signifies the most similar, while 4 indicates the most dissimilar.
We use the image - caption pairs in Appendix~\ref{sec:modified_dataset} to create
11,000 images from the ImageNet~\cite{5206848} validation set and introduced random modifications by SDEdit~\cite{meng2021sdedit}, InstructPix2pix~\cite{brooks2023instructpix2pix}. Resource constraints limited us to employing five human annotators, with each annotator labeling every datum. The entire labeling process incurred a cost of \$600.

\begin{figure}[ht]
    \centering
    \includegraphics[width=0.68\linewidth]{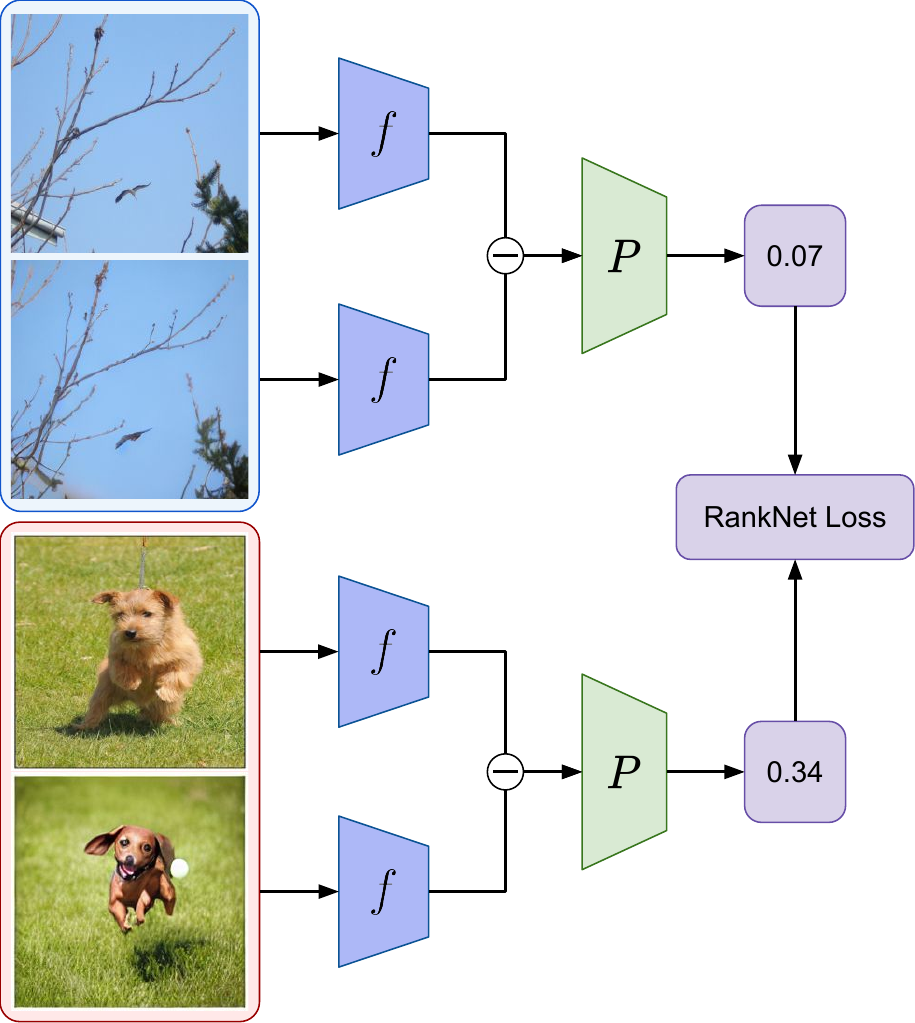}
    \vspace{-0.5em}
    \caption{Illustration of the Siamese Network processing flow for image pairs using RankNet Loss. Each image within a pair undergoes transformation using the ResNet-50 backbone, denoted as $f$. The resultant features are then subtracted and channeled through an MLP $P$, yielding a similarity score. This score subsequently informs the computation of the RankNet Loss.}
    \vspace{-1em}
    \label{fig:ranknet}
\end{figure}

\noindent\textbf{Training Details of $\SimName$ Siamese Network.} 
\addcontentsline{toc}{subsection}{Training Details of $\SimName$ Siamese Network.}
Recognizing that image similarity inherently involves comparing pairs of images, we utilized a neural architecture ideally suited for this scenario: the Siamese Network~\cite{bromley1993signature}. The Siamese network structure is specifically tailored for tasks like image similarity, where it processes two input images through shared weights to derive a similarity measure between them. In our implementation, we adopted ResNet-50~\cite{he2016deep} as the backbone of the Siamese network.
Prior to training, it became essential to convert human rankings into scores. We achieved this by normalizing the rank within its maximum rank and then averaging over all images:

\begin{equation}
    \text {Score}_i=\frac{1}{n}\sum_{j=1}^n\frac{\text{Rank}_j}{\max(\text{Rank})}.
\end{equation}
For every training iteration, two image pairs were randomly sampled from a 5-image tuple. A label was assigned with a value of 1 if the first image pair had a higher mean ranking compared to the second, otherwise, it was labeled 0. Subsequently, we utilized the RankNet Loss~\cite{burges2005learning}, with $i$ and $j$ denoting distinct pairs within each sample and $y_{ij}$ representing the binary label derived from human annotations:

\begin{equation}
    P_{ij} = \frac{1}{1 + e^{-(s_i - s_j)}} \label{eq:ranknet_prob},
\end{equation}
where the pairwise RankNet loss is:
\begin{equation}
    L(y_{ij}, P_{ij}) = - y_{ij} \log(P_{ij}) - (1 - y_{ij}) \log(1 - P_{ij}) \label{eq:ranknet_loss}.
\end{equation}
This process is illustrated in Fig.~\ref{fig:ranknet}.

\noindent\textbf{Evaluation of the learnt $\SimName$ Score.}
\addcontentsline{toc}{subsection}{Evaluation of the learnt $\SimName$ Score.}
In our evaluation, we use Spearman Distance as the primary metric to assess the dissimilarity in rankings of 5-image tuples, with a focus on determining how closely each method aligns with human judgment. A lower Spearman Distance value indicates a closer alignment to the reference ranking, thus better mirroring human perception. According to the results detailed in Table~\ref{tab:viseval}, $\SimName$ closely approximates human rankings with a Spearman Distance of 2.89 (at a similar level of the human cross-validation scores to a held-out human annotator, Table~\ref{tab:viseval}). This contrasts with traditional metrics like LPIPS and SSIM, which show higher disparities in their Spearman Distance scores. Moreover, our analysis revealed significant variability in human judgment across different annotators, highlighting the subjective nature of visual assessments and the complexity involved in developing algorithms that accurately reflect human perception.

\vspace{-.5em}
\section{Additional Results}
\vspace{-.5em}
\subsection{Additional Results on InstructPix2pix}
\vspace{-.5em}
\label{append:diffdataset}

\begin{table*}[t!]
\centering
\vspace{-.5em}
\resizebox{1.0\textwidth}{!}{
\begin{tabular}{c|cc|cc|cc|cc|cc|cc}
\hline
                          & \multicolumn{2}{c|}{\textbf{SDEdit\cite{meng2021sdedit}}}                                                                & \multicolumn{2}{c|}{\textbf{InstructPix2Pix\cite{brooks2023instructpix2pix}}}                                                         & \multicolumn{2}{c|}{\textbf{Zero 1-to-3\cite{liu2023zero}}}                                                          & \multicolumn{2}{c|}{\textbf{InPaint\cite{lugmayr2022repaint}}}                                                               & \multicolumn{2}{c|}{\textbf{DALL·E 2\cite{ramesh2022hierarchical}}} & \multicolumn{2}{c}{\textbf{Average}}                                                              \\ \hline
                          & \multicolumn{1}{c}{\textbf{\begin{tabular}[c]{@{}c@{}}TPR ($\uparrow$) @\\ 1\%FPR\end{tabular}}} & \textbf{AUC ($\uparrow$)}   & \multicolumn{1}{c}{\textbf{\begin{tabular}[c]{@{}c@{}}TPR ($\uparrow$) @\\ 1\%FPR\end{tabular}}} & \textbf{AUC ($\uparrow$)}   & \multicolumn{1}{c}{\textbf{\begin{tabular}[c]{@{}c@{}}TPR ($\uparrow$) @\\ 1\%FPR\end{tabular}}} & \textbf{AUC ($\uparrow$)}  & \multicolumn{1}{c}{\textbf{\begin{tabular}[c]{@{}c@{}}TPR ($\uparrow$) @\\ 1\%FPR\end{tabular}}} & \textbf{AUC ($\uparrow$)}   & \multicolumn{1}{c}{\textbf{\begin{tabular}[c]{@{}c@{}}TPR ($\uparrow$) @\\ 1\%FPR\end{tabular}}} & \textbf{AUC ($\uparrow$)} & \multicolumn{1}{c}{\textbf{\begin{tabular}[c]{@{}c@{}}TPR ($\uparrow$) @\\ 1\%FPR\end{tabular}}} & \textbf{AUC ($\uparrow$)}  \\ \hline
\textbf{\begin{tabular}[c]{@{}c@{}}DwtDctSVD\\ \cite{4554423}\end{tabular}}        & \multicolumn{1}{c}{\large 0.035}                                                         & \large 0.560          & \multicolumn{1}{c}{\large 0.040}                                                          & \large 0.530          & \multicolumn{1}{c}{\large 0.025}                                                         & \large 0.525         & \multicolumn{1}{c}{\large 0.020}                                                         & \large 0.600          & \multicolumn{1}{c}{\large 0.015}                                                         & \large 0.520          & \multicolumn{1}{c}{\large 0.027}                                                & \large 0.547\\ \hline
\textbf{\begin{tabular}[c]{@{}c@{}}HiDDeN\\ \cite{zhu2018hidden}\end{tabular}}           & \multicolumn{1}{c}{\large 0.030}                                                         & \large 0.540          & \multicolumn{1}{c}{\large 0.050}                                                         & \large 0.630          & \multicolumn{1}{c}{\large 0.015}                                                        & \large 0.610         & \multicolumn{1}{c}{\large 0.040}                                                         & \large 0.700          & \multicolumn{1}{c}{\large 0.005}                                                         & \large 0.600          & \multicolumn{1}{c}{\large 0.028}                                               & \large 0.616\\ \hline
\textbf{\begin{tabular}[c]{@{}c@{}}Stable Sig*\\ \cite{fernandez2023stable}\end{tabular}} & \multicolumn{1}{c}{\large 0.015}                                                         & \large 0.520          & \multicolumn{1}{c}{\large 0.020}                                                          & \large 0.600          & \large 0.005 & \large 0.515                                                                           & \multicolumn{1}{c}{\large 0.040}                                                         & \large 0.640          & \multicolumn{1}{c}{\large 0.015}                                                         & \large 0.570          & \multicolumn{1}{c}{\large 0.019}                                                & \large 0.569\\ \hline
\textbf{\begin{tabular}[c]{@{}c@{}}Tree-Ring*\\\cite{wen2023tree} \end{tabular}}        & \multicolumn{1}{c}{\large 0.160}                                                         & \large 0.890          & \multicolumn{1}{c}{\large 0.150}                                                         & \large 0.840          & \large 0.120 &\large 0.810                                                                            & \multicolumn{1}{c}{\large 0.230}                                                         & \large 0.862          & \multicolumn{1}{c}{\large 0.250}                                                         & \large 0.870          & \multicolumn{1}{c}{\large 0.182}                                                & \large 0.854\\ \hline
\textbf{\begin{tabular}[c]{@{}c@{}}$\AlgName$\\ (Ours)\end{tabular}}             & \multicolumn{1}{c}{\textbf{\large 0.915}}                                                & \textbf{\large 0.981} & \multicolumn{1}{c}{\textbf{\large 0.947}}                                                & \textbf{\large 0.988} & \multicolumn{1}{c}{\textbf{\large 0.881}}                                                 & \textbf{\large 0.981} & \multicolumn{1}{c}{\textbf{\large 0.885}}                                                & \textbf{\large 0.980} & \multicolumn{1}{c}{\textbf{\large 0.723}}                                                & \textbf{\large 0.937} & \multicolumn{1}{c}{\textbf{\large 0.870}}                                                & \textbf{\large 0.973}\\ \hline
\end{tabular}

}
\caption{Comparison of watermark detection under various diffusion image perturbations,\textbf{ InstructPix2Pix} samples. *For Stable Sig and Tree-Ring, results are derived from their synthetic data generated using textual guidance from our evaluation datasets (image-text pairs).}
\vspace{-2em}
\label{tab:laioneval}
\end{table*}
Considering the penitential data leakage, we prioritize using a custom ImageNet~\cite{5206848} dataset for evaluations in Section~\ref{sec:mainevaluation}. To facilitate a comprehensive analysis, we also include the evaluation results over the InstructPix2Pix~\cite{brooks2023instructpix2pix} (based on LAION-5B). 
Rather than leveraging visual language models to generate image captions, InstructPix2Pix first collected 700 images from LAION along with human-written captions. Humans then provided editing instructions for each image-caption pair. Using these caption-instruction examples, the authors fine-tuned GPT-3 models~\cite{brown2020language} to automatically generate additional editing instructions for LAION images (a total of more than 0.4 million image-text pairs). Similar to the main evaluation (using ImageNet, Section \ref{sec:case2exp}), we randomly select 2,000 image-instruction pairs from the InstructPix2Pix dataset to assess the performance of $\AlgName$ and other baseline methods under various diffusion model image perturbations. We employ a $\SimName$ range of 0.3-0.5 to ensure that the perturbation strength remains within the range perceptible to humans. The results, presented in Table~\ref{tab:laioneval}, largely align with those from our main evaluation, demonstrating the robustness of our method across multiple datasets.

\subsection{Type-2 Perturbations In-depth Case Study}
We now conduct a case study of $\AlgName$ under varied diffusion perturbation conditions. This includes assessing the impact of different Stable Diffusion versions (of the SDEdit) and the effects under iterative image modifications.
\begin{figure}[h]
    \centering
    \vspace{-1em}
    \includegraphics[width=0.98\linewidth]{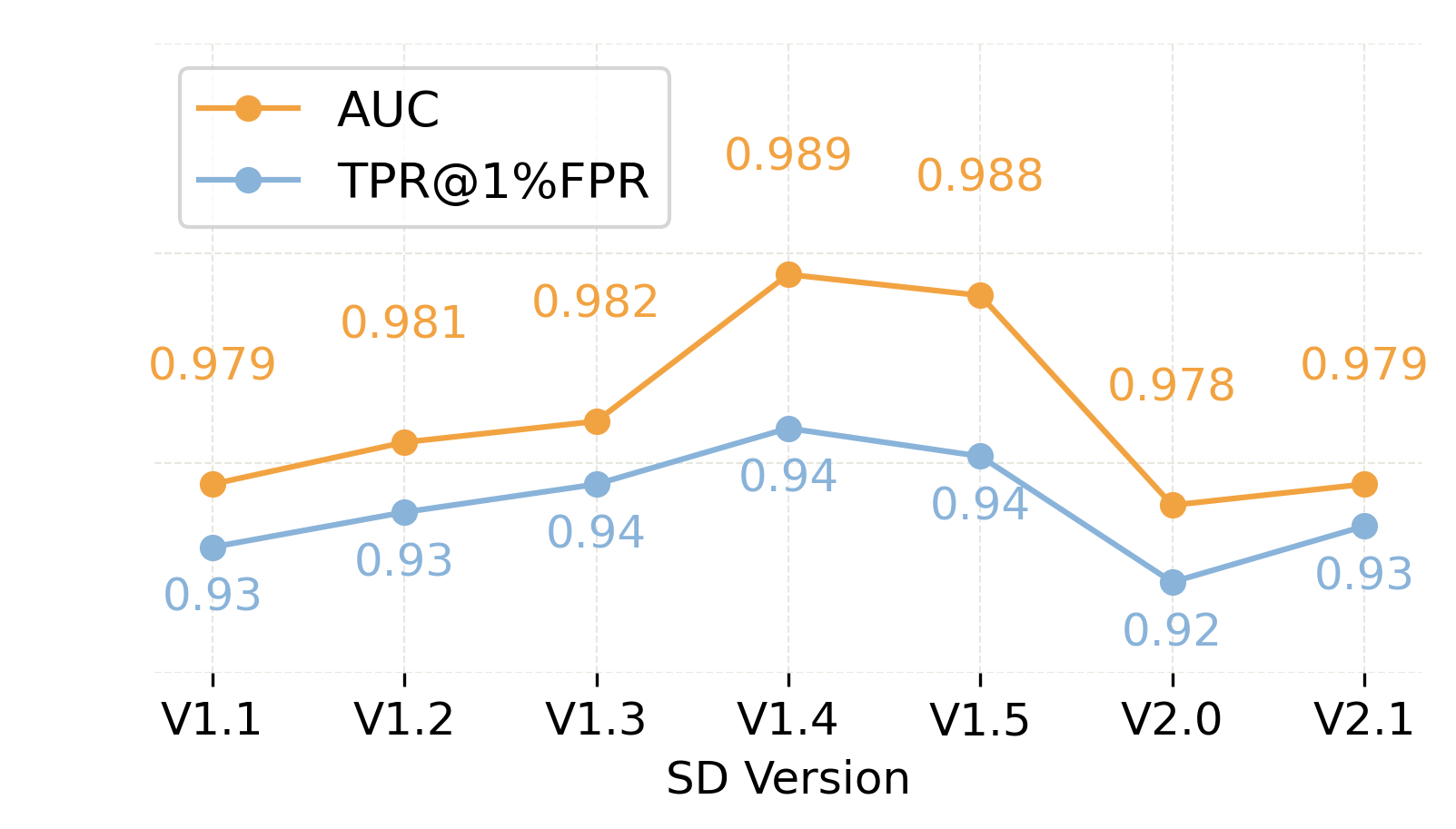}
    \vspace{-.5em}
    \caption{Use different stable diffusion version as the base model of SDEdit~\cite{meng2021sdedit} for evaluation (training uses V1.4 for the SDEdit).}
    \label{fig:diffv}
    \vspace{-1em}
\end{figure}

\noindent\textbf{Different SD Version.} 
\addcontentsline{toc}{subsection}{\qquad Different SD Version.} 
We additionally evaluated how different versions of Stable Diffusion may impact $\AlgName$'s detection capability, as presented in Fig.~\ref{fig:diffv} . Using Stable Diffusion versions 1.1, 1.2, 1.3, 1.4 (the one incoporated during $\AlgName$ training), 1.5, 2.0 and 2.1 to generate image perturbations at an $\SimName$ range of 0.3 to 0.5, we tested $\AlgName$'s resilience across these model updates. Our experiments showed that the choice of Stable Diffusion version introduces only minor variation in both AUC score and True Positive Rate at 1\% False Positive Rate. The small variance range underscores $\AlgName$'s consistent robustness across Stable Diffusion versions. The minimal impact from model updates highlights that $\AlgName$ effectively generalizes against diffusion perturbations without overfitting to any specific version. By maintaining steady performance despite changes in the perturbation model, $\AlgName$ demonstrates its capability to handle diffusion image edits in a version-agnostic manner.

\begin{figure}[h]
    \centering
    \includegraphics[width=0.98\linewidth]{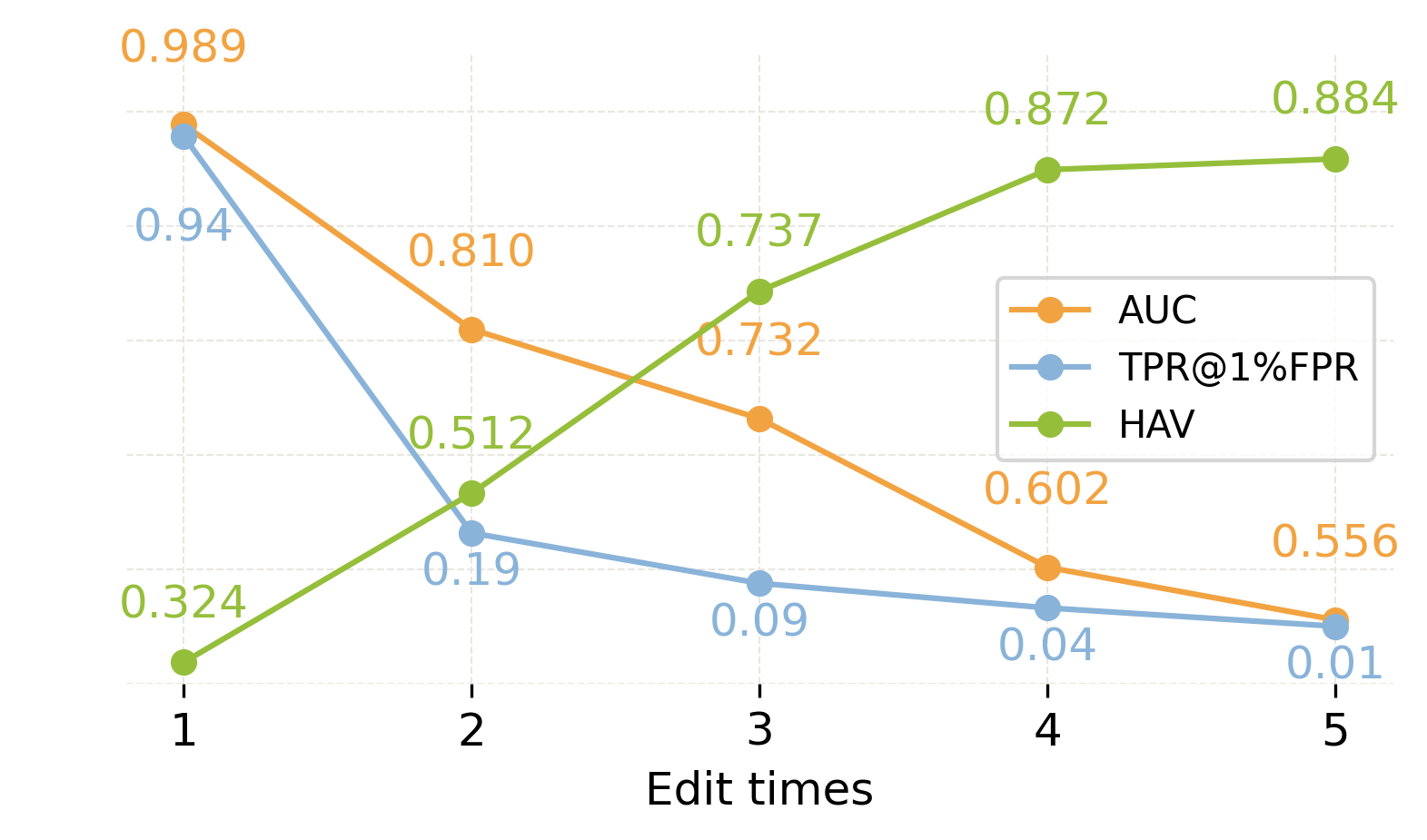}
    \vspace{-.5em}
    \caption{Iterative image variation via SDEdit~\cite{meng2021sdedit} simulating cases of iterative modifications.}
    \vspace{-1em}
    \label{fig:edittimes}
\end{figure}
\noindent\textbf{Iterative Perturbations.} 
\addcontentsline{toc}{subsection}{\qquad Iterative Perturbations.} 
We also analyzed how applying iterative perturbations impacts $\AlgName$’s detection performance in Fig.~\ref{fig:edittimes}. Specifically, we tested perturbations involving 1 to 5 sequential applications of SDEdit~\cite{meng2021sdedit} on the images. Our experiments revealed that with increased perturbation rounds, both the AUC score and True Positive Rate at 1\% False Positive Rate decay rapidly. After the first round of SDEdit edits with an AUC of 0.989 and TPR of 0.945, just two additional rounds drop the metrics to 0.810 AUC and 0.188 TPR. By the fifth round of perturbations, the AUC declines to 0.556 and TPR to 0.010. Concurrently, $\SimName$ also increased rapidly, surpassing 0.5 after only two perturbations, which is beyond our defined range and may not reflect enough information derivations, thus out of the evaluation scope.
This trade-off underscores that while $\AlgName$ demonstrates significant resilience to perturbations within the defined range, its effectiveness diminishes beyond this threshold. However, within the acceptable range of perturbations, our method has shown to be robust, effectively maintaining high detection accuracy and demonstrating its practicality in real-world scenarios.

\subsection{Additional Evaluation \& Results}
\label{append:more_results}

\begin{table*}[t!]
\centering
\resizebox{1.0\textwidth}{!}{
\begin{tabular}{c|cc|cc|cc|cc|cc}
\hline
                          & \multicolumn{2}{c|}{\textbf{Pix2Pix\cite{isola2017image}}}                                                               & \multicolumn{2}{c|}{\textbf{CycleGAN\cite{zhu2017unpaired}}}                                                          & \multicolumn{2}{c|}{\textbf{CUT\cite{park2020contrastive}}}                                                          & \multicolumn{2}{c|}{\textbf{LaMa\cite{suvorov2022resolution}}}                                                               & \multicolumn{2}{c}{\textbf{Average}}                                                              \\ \hline
                          & \multicolumn{1}{c}{\textbf{\begin{tabular}[c]{@{}c@{}}TPR ($\uparrow$) @\\ 1\%FPR\end{tabular}}} & \textbf{AUC ($\uparrow$)}   & \multicolumn{1}{c}{\textbf{\begin{tabular}[c]{@{}c@{}}TPR ($\uparrow$) @\\ 1\%FPR\end{tabular}}} & \textbf{AUC ($\uparrow$)}   & \multicolumn{1}{c}{\textbf{\begin{tabular}[c]{@{}c@{}}TPR ($\uparrow$) @\\ 1\%FPR\end{tabular}}} & \textbf{AUC ($\uparrow$)}  & \multicolumn{1}{c}{\textbf{\begin{tabular}[c]{@{}c@{}}TPR ($\uparrow$) @\\ 1\%FPR\end{tabular}}} & \textbf{AUC ($\uparrow$)}   & \multicolumn{1}{c}{\textbf{\begin{tabular}[c]{@{}c@{}}TPR ($\uparrow$) @\\ 1\%FPR\end{tabular}}} & \textbf{AUC ($\uparrow$)}  \\ \hline
\textbf{\begin{tabular}[c]{@{}c@{}}DwtDctSVD\\ \cite{4554423}\end{tabular}}        & \multicolumn{1}{c}{\large 0.042}                                                         & \large 0.572          & \multicolumn{1}{c}{\large 0.037}                                                          & \large 0.552          & \multicolumn{1}{c}{\large 0.034}                                                         & \large 0.544         & \multicolumn{1}{c}{\large 0.029}                                                         & \large 0.567          & \multicolumn{1}{c}{\large 0.036}                                                & \large 0.559\\ \hline
\textbf{\begin{tabular}[c]{@{}c@{}}HiDDeN\\ \cite{zhu2018hidden}\end{tabular}}           & \multicolumn{1}{c}{\large 0.054}                                                         & \large 0.563          & \multicolumn{1}{c}{\large 0.061}                                                         & \large 0.642          & \multicolumn{1}{c}{\large 0.042}                                                        & \large 0.617         & \multicolumn{1}{c}{\large 0.038}                                                         & \large 0.708          & \multicolumn{1}{c}{\large 0.049}                                               & \large 0.633\\ \hline
\textbf{\begin{tabular}[c]{@{}c@{}}Stable Sig*\\ \cite{fernandez2023stable}\end{tabular}} & \multicolumn{1}{c}{\large 0.051}                                                         & \large 0.538          & \multicolumn{1}{c}{\large 0.048}                                                          & \large 0.618          & \large 0.026 & \large 0.528                                                                           & \multicolumn{1}{c}{\large 0.033}                                                         & \large 0.657          & \multicolumn{1}{c}{\large 0.040}                                                & \large 0.585\\ \hline
\textbf{\begin{tabular}[c]{@{}c@{}}Tree-Ring*\\\cite{wen2023tree} \end{tabular}}        & \multicolumn{1}{c}{\large 0.038}                                                         & \large 0.543          & \multicolumn{1}{c}{\large 0.017}                                                         & \large 0.511          & \large 0.015 &\large 0.518                                                                            & \multicolumn{1}{c}{\large 0.175}                                                         & \large 0.800          & \multicolumn{1}{c}{\large 0.061}                                                & \large 0.593\\ \hline
\textbf{\begin{tabular}[c]{@{}c@{}}$\AlgName$\\ (Ours)\end{tabular}}             & \multicolumn{1}{c}{\textbf{\large 0.163}}                                                & \textbf{\large 0.862} & \multicolumn{1}{c}{\textbf{\large 0.214}}                                                & \textbf{\large 0.871} & \multicolumn{1}{c}{\textbf{\large 0.186}}                                                 & \textbf{\large 0.886} & \multicolumn{1}{c}{\textbf{\large 0.155}}                                                & \textbf{\large 0.831} & \multicolumn{1}{c}{\textbf{\large 0.180}}                                                & \textbf{\large 0.863}\\ \hline
\end{tabular}
}
\caption{Comparison of watermark detection under various image editingtechniques with \textbf{GANs}. *For Stable Sig and Tree-Ring, results are derived from their synthetic data generated using textual guidance from our evaluation datasets (image-text pairs).}
\vspace{-2.5em}
\label{tab:GANeval}
\end{table*}

\begin{figure*}[h]
    \centering
    \includegraphics[width=0.68\linewidth]{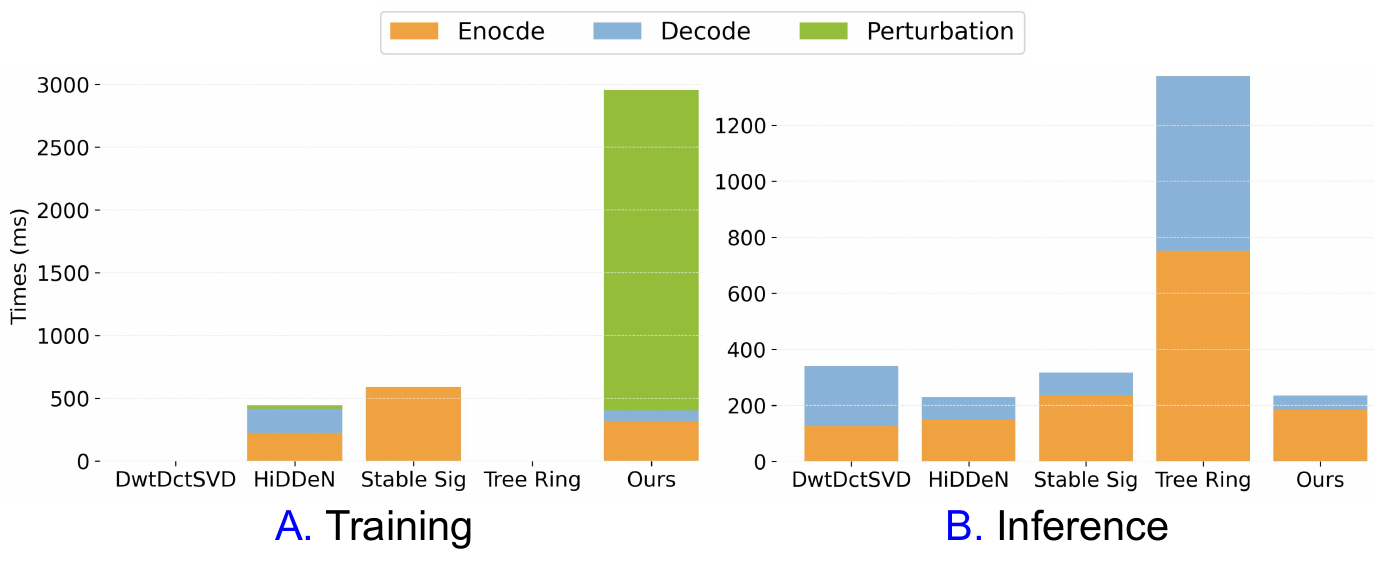}
    \vspace{-1em}
    \caption{Time overhead analysis of watermarking methods for \textcolor{blue}{\textbf{A}}. Training time and \textcolor{blue}{\textbf{B}}. Inference time.}
    \vspace{-1.8em}
    \label{fig:timeoverhead}
\end{figure*}

\noindent\textbf{Evaluation on GAN-based image variations.}
\addcontentsline{toc}{subsection}{\qquad Evaluation on GAN-based image variations.}
In addition to diffusion models, Generative Adversarial Networks (GANs) represent a significant category in the image editing domain, capable of functions similar to diffusion models. For example, Pix2pix~\cite{isola2017image} enables image-to-image translation through conditional adversarial networks; CycleGAN~\cite{zhu2017unpaired} facilitates unpaired image translation with adversarial and cycle consistency losses for style transfer; CUT~\cite{park2020contrastive} employs contrastive learning for one-sided unpaired image translation; and LaMa~\cite{suvorov2022resolution} specializes in high-resolution image inpainting using fast Fourier convolutions and a high receptive field loss. We evaluated these methods in a zero-shot manner as perturbations and presented $\AlgName$ alongside other baseline watermarking techniques in Table~\ref{tab:GANeval}.
Hidden, Stable Signature, and DctDwtSvd exhibit the same low AUC scores as the main evaluation. Tree-Ring, which embeds its hand-crafted watermark in the diffusion latent space, is significantly impacted by GAN-based modifications. This is due to the differences between the latent spaces of diffusion models and GANs, where in several methods, its AUC approaches 0.5 (random guessing), a drop of over 0.3 compared to diffusion model perturbations.
$\AlgName$ outperforms other methods across different GAN models, maintaining the highest AUC values, averaging 0.863. This indicates better robustness of $\AlgName$ to image modifications induced by GANs, effectively detecting watermarks even under significant alterations. However, it is noteworthy that $\AlgName$ shows a performance decrease compared to its effectiveness under diffusion model image editing. This reduction can be attributed to the fundamental differences in the generation processes of GANs, which are not incorporated in the training process of $\AlgName$. Given that GAN-based methods are not the primary focus of this paper and the current landscape of image variations, we limit our discussion to this zero-shot evaluation.


\noindent\textbf{Adaption on DALL$\cdot$E 2.}
\label{append:dalle2}
\addcontentsline{toc}{subsection}{\qquad Adaption on DALL$\cdot$E 2.}
DALL$\cdot$E 2 currently offers its API solely for black-box image variation purposes, meaning users can only upload images and receive the edited versions in return. As highlighted in Section~\ref{sec:case2exp}, our zero-shot evaluation on DALL$\cdot$E 2 yielded an AUC of 0.934, which is commendable but not optimal. To further demonstrate $\AlgName$'s adaptability to unseen and black-box perturbations, we performed fine-tuning on such perturbations. Utilizing the same dataset as in our main evaluation, we opted for a small batch size of 10 and updated the model for 600 steps. This fine-tuning process successfully increased the AUC to 0.980 and the TPR at 1\% FPR to 0.824, at a cost of only \$96. This adaptation demonstrates that our method's efficacy can be improved via few-shot fine-tuning for unseen perturbations.

\noindent\textbf{Quantify Watermark Stealthiness.}
\addcontentsline{toc}{subsection}{\qquad Quantify Watermark Stealthiness.}
Table~\ref{tab:vis_diff} evaluate watermarking techniques based on PSNR, SSIM~\cite{wang2004image}, and LPIPS metrics~\cite{zhang2018unreasonable}. 
We omit the comparison to diffusion-model-centric watermarking methods as they cannot be adopted to watermark a given image.
DwtDctSVD scores the highest in PSNR (32.2197), indicating minimal pixel-based image differences. However, it's worth noting that a higher PSNR doesn't always correlate to perceived visual similarity due to the non-linear nature of human visual perception. All methods show similar SSIM values, implying consistent structural integrity, with our method registering the least structural degradation (SSIM = 0.89372). Notably, our method excels in the LPIPS metric (0.06572), outperforming others by 40-50\%. This suggests our watermarks are perceptually less noticeable, better preserving the original image's visual quality.

\vspace{-1em}
\begin{table}[h]
\centering
\resizebox{0.45\linewidth}{!}{
\begin{tabular}{c|c|c|c}
\hline
\textbf{}          & \textbf{PSNR↑}   & \textbf{SSIM↓}   & \textbf{LPIPS↓} \\ \hline
\textbf{DwtDctSVD} & \textbf{32.2197} & 0.89598          & 0.10785                              \\ \hline
\textbf{HiDDeN}    & 30.8405          & 0.89548 & 0.12753                              \\ \hline
\textbf{Ours}      & 30.1354          & \textbf{0.89372}          & \textbf{0.06572}                     \\ \hline
\end{tabular}}
\caption{Comparison of watermarking methods using visual similarity metrics. The best results for each metric are highlighted in bold.}
\vspace{-2.5em}
\label{tab:vis_diff}
\end{table}

\noindent\textbf{Training Overhead Analysis.}
\addcontentsline{toc}{subsection}{\qquad Training Overhead Analysis.}
We present our training and inference time analysis in Fig.~\ref{fig:timeoverhead}. All evaluations were performed on a server equipped with 2 $\times$ AMD EPYC 7736 CPUs and 8 $\times$ Nvidia Tesla A100 GPUs. 

In the training overhead analysis, DwtDctSCD~\cite{4554423} and Tree-Ring~\cite{wen2023tree} embed watermarks directly without an optimization phase, eliminating the need for training time. When considering only the watermark components, encoder, and decoder, our approach becomes the most efficient. The Stable Signature~\cite{fernandez2023stable} requires fine-tuning the latent decoder over a trained watermark decoder. Since the latent decoder is relatively larger, this fine-tuning demands substantial training overhead. However, our method is the only one that can incorporate diffusion model perturbation into the training process. While the diffusion step introduces significant training time, embedding a personalized key allows a trained watermark model to be directly deployed to multiple users, mitigating the impact of training overhead.

In terms of inference overhead, DwtDctSVD embeds the watermark post-matrix decomposition, a process exclusively performed on the CPU. This limitation results in increased inference time, even on high-performance server CPUs. Conversely, the Tree-Ring method requires the use of a diffusion model to reverse the diffusion step, transforming the image back into the diffusion latent space, which significantly increases the time overhead. It is important to consider that in real-world applications, watermark agents frequently query and decode images from the internet to determine if they contain watermarks. Therefore, decoding processes occur more frequently than encoding. This frequent need for decoding emphasizes the importance of efficiency in the decoding process, making it a more critical factor than encoding efficiency in practical scenarios. Our method, through a specially designed decoder, achieves the lowest decoding time.

\begin{table}[h]
\centering
\vspace{-.5em}
\resizebox{.6\linewidth}{!}{
\begin{tabular}{c|c|c|c|c}
\hline
      & \begin{tabular}[c]{@{}c@{}}Full-Model\\ Retraing\end{tabular} & \begin{tabular}[c]{@{}c@{}}Fine-Tune\\ 100 steps\end{tabular} & \begin{tabular}[c]{@{}c@{}}Fine-Tune\\ 1000 steps\end{tabular} & \begin{tabular}[c]{@{}c@{}}Fine-Tune\\ 5000 steps\end{tabular} \\ \hline
Time$\downarrow$  & 40h                                                           & 0.16h                                                         & 1.6h                                                           & 8h                                                             \\ \hline
E-TPR$\downarrow$ & 0.075                                                         & 0.164                                                         & 0.107                                                          & 0.083                                                          \\ \hline
\end{tabular}}
\caption{Training time for an 8$\times$A100 GPU server, with E-TPR indicating the misclassified watermarking cases between the original and fine-tuned models.}
\vspace{-2.5em}
\label{tab:higherror}
\end{table}

Regarding inference, our method achieves the lowest overhead owing to its lightweight encoder and decoder design. In contrast, the Tree-Ring method, despite its performance being second only to $\AlgName$, necessitates diffusion latent inversion for each watermark embedding and detection, leading to significant computational overhead. A key advantage of our method is its capability for zero-shot watermark embedding, which means a single pre-trained model suffices for multiple applications, thereby mitigating the impact of training overhead.

\subsection{Mismatch Analysis (False Positive Case Study)}
\label{append: personalkey}

\begin{figure}[h]
    \centering
    \includegraphics[width=0.9\linewidth]{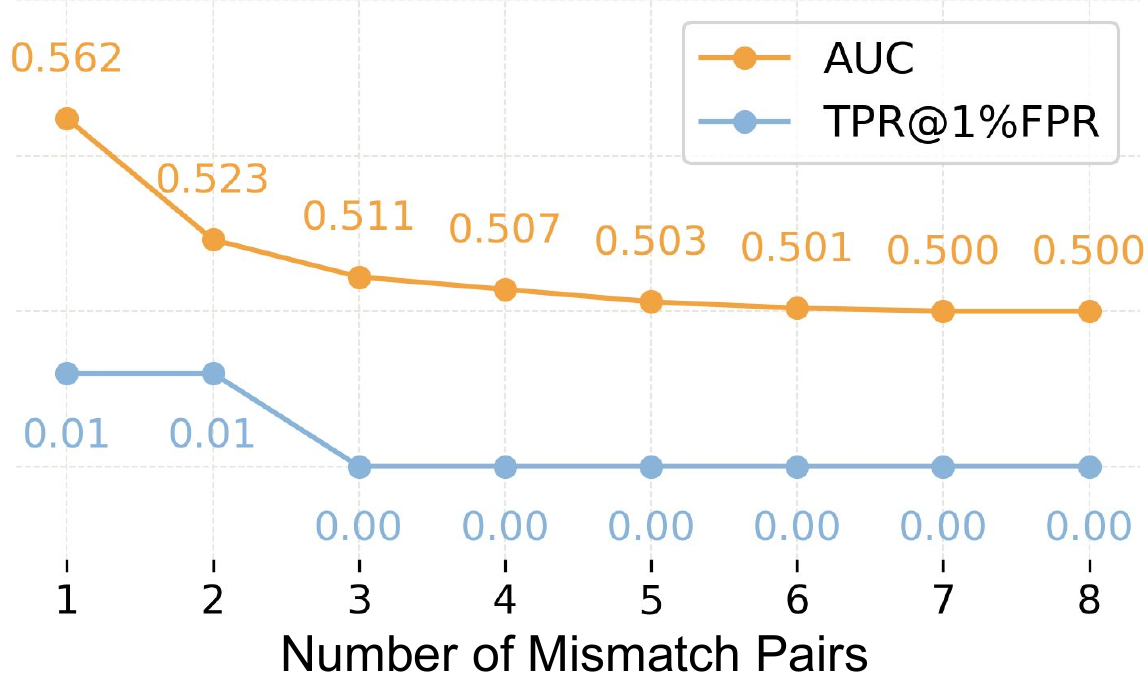}
    \vspace{-0.5em}
    \caption{Detection performance in the presence of mismatched Jigsaw patch pairs. The x-axis represents the number of mismatched patch pairs, with a range from a minimum of 1 pair (two patches swapped) to a maximum of 8 pairs (all 16 patches flipped).}
    \vspace{-1em}
    \label{fig:mismatch}
\end{figure}

As mentioned in Section~\ref{sec:threatmodels}, evaluating the False Positive Rate (FPR) is crucial to ensure that the secret keys of other users are not misclassified as those of the target user. Our method employs a Jigsaw combination order as the watermark key. Consider an extreme case where two users generate very similar Jigsaw combination orders, differing only by the swapping of two Jigsaw patch pairs, while the rest remain identical. To rigorously assess our method's FPR in such extreme cases, we conducted experiments and present the results in Fig.~\ref{fig:mismatch}. In this extreme scenario, with just one mismatched patch pair (two patches swapped), we observed an AUC of 0.56 and a TPR at a 1\% FPR of only 0.01. These scores, closely approximating the random guessing baseline of 0.5, suggest that our method effectively minimizes false positives even under highly similar Jigsaw combination orders, thereby affirming its reliability in distinguishing between different user keys.
The Jigsaw methodology offers a significant improvement over the idea of retraining or fine-tuning, as evidenced by our results in Table~\ref{tab:higherror}. This innovative approach enables the watermarking system to be both adaptable and scalable.

\vspace{-1em}
\section{Methodology Details}
\vspace{-.5em}

\subsection{Training Algorithm Details}
\label{sec:jigmarkalg}

\addcontentsline{toc}{subsection}{$\AlgName$ Training Algorithm.}
\begin{algorithm}[h]
   \caption{Pseudo-code of $\AlgName$ Training}
   \label{algorithm:pythoncode}
\begin{PythonB}
# Initialize watermark Encoder and Decoder
E, D = encoder(), decoder()
opt_e, opt_d = Adamw(E, D)

for x, inst in dataloader:
    # Change the information into shuffle order
    Si, S = random_shffle()
    # Encode input data
    x_w = Si(E(S(x)))
    # Apply perturbation P with to the data
    x_p, x_w_p = P((x, x_w), inst)
    # Decode the original and perturbed data
    i, i_w, i_p, i_w_p = D(S(x, x_w, x_p, x_w_p))
    # Decode the random shuffle data
    S_r = random_shffle()
    r_i_w, r_i_w_p = D(S_r(x_w, x_w_p))
    # Positive pair and negative pair
    pos = cat(i_w, i_w_p)
    neg = cat(I, i_p, r_i_w, r_i_w_p)
    
    # Compute visual loss
    L_v = visual_loss(x, x_w)
    # Compute watermark loss
    L_w = wm_loss(pos, neg)
    # Compute total loss 
    total_loss = L_w + l_v
    
    # Backpropagate the loss
    total_loss.backward()
    # Update the parameters
    opt_e.step(), opt_d.step()
\end{PythonB}
\vspace{-.5em}
\end{algorithm}

We provide the detailed training algorithm of $\AlgName$ in Algorithm \ref{algorithm:pythoncode}. The explanations of the key components and detailed workflow is presented in the main text, Section \ref{sec:methodology}.

\subsection{Additional Engineering Details}

\noindent\textbf{Enhancing Watermark Detection.} 
For effective watermark detection, our decoder $D$ differentiates easily between watermarks in standard images $x$ and $x_w$ and those in perturbed versions $x'$ and $x'_w$. To improve its performance with perturbed images, where distortions obscure watermarks, we generate three distinct perturbed instances for each original and watermarked image pair. This method, utilizing varied prompts, equips $D$ to handle diverse alteration scenarios, ensuring robust watermark detection across a range of image conditions. This approach is a pivotal aspect of our implementation for consistent watermark identification.

\noindent\textbf{Gradient Clipping.}
In our training process, certain generated images may exhibit distortions or become unreadable, leading to unstable gradients that can compromise the training process of both the encoder and decoder. To enhance training stability, we incorporate an advanced gradient clipping technique, AutoClip~\cite{seetharaman2020autoclip}. AutoClip employs a history-based approach, utilizing the percentage of past gradient norms to determine an optimal clipping threshold. For our implementation, we adhere to the original paper's guidelines and set the clipping threshold at 10 percent. This strategic application of gradient clipping significantly stabilizes the training, ensuring smoother optimization and mitigating issues caused by distorted or unreadable image generations.

\noindent\textbf{Replace BN with GN.}
\quad Image perturbations can drastically alter an image's statistics. For instance, changes in brightness directly modify pixel values, consequently altering the image's mean. Concurrently, the extensive size of the diffusion model restricts the training batch size. This combination of factors adversely impacts the performance of Batch Normalization (BN)~\cite{ba2016layer}. So we replace the Batch Normalization (BN) with Group Normalization (GN)~\cite{wu2018group}. GN operates by normalizing groups of channels, eliminating the need for large batch sizes. This ensures stable and consistent training, even when image statistics vary widely. As the ConvNeXt block do not have BN layer, we only replace the BN in the decoder.

\begin{figure}[h]
    \centering
    \includegraphics[width=0.98\linewidth]{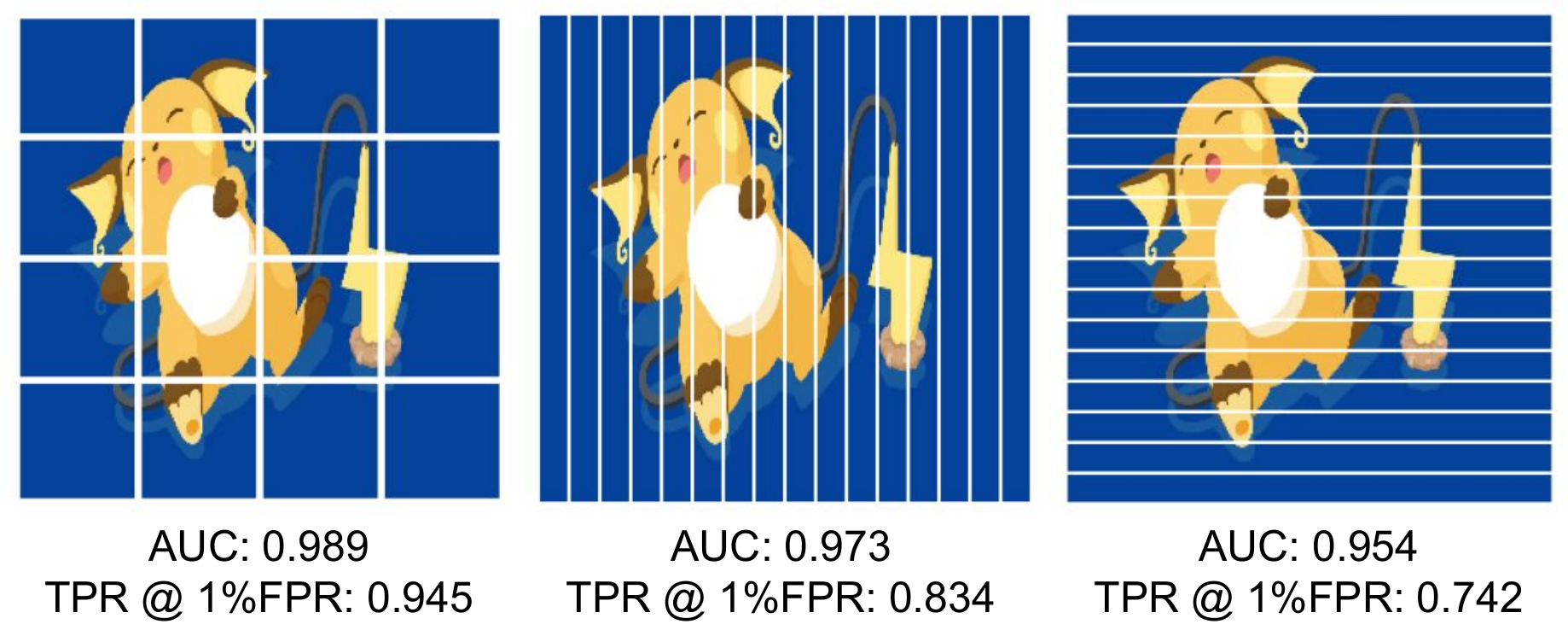}
    \vspace{-.5em}
    \caption{Performance with different Jigsaw piece shapes.}
    \vspace{-1em}
    \label{fig:diffshape}
\end{figure}

\noindent\textbf{Different Jigsaw Shape.}
Additionally, we studied how the shape of the image segmentation used in $\AlgName$'s impacts robustness. We tested square blocks, vertical rectangular strips, and horizontal rectangular strips, as show in Fig.~\ref{fig:diffshape}. The results show that square blocks achieve the highest detection metrics, with an AUC of 0.989 and True Positive Rate at 1\% False Positive Rate of 0.945. Vertical rectangular strips lead to a minor drop in AUC to 0.973 and TPR to 0.834. Horizontal rectangular strips result in the lowest scores of 0.954 AUC and 0.742 TPR. This variance indicates that square blocks, providing a more balanced segmentation, are optimal for embedding robust watermarks. We posit the greater dimensionality of square blocks (along both image axes) facilities more the encoder learning more information relate to the original image semantic.

\begin{figure}[h]
    \centering
    \vspace{-1em}
    \includegraphics[width=0.95\linewidth]{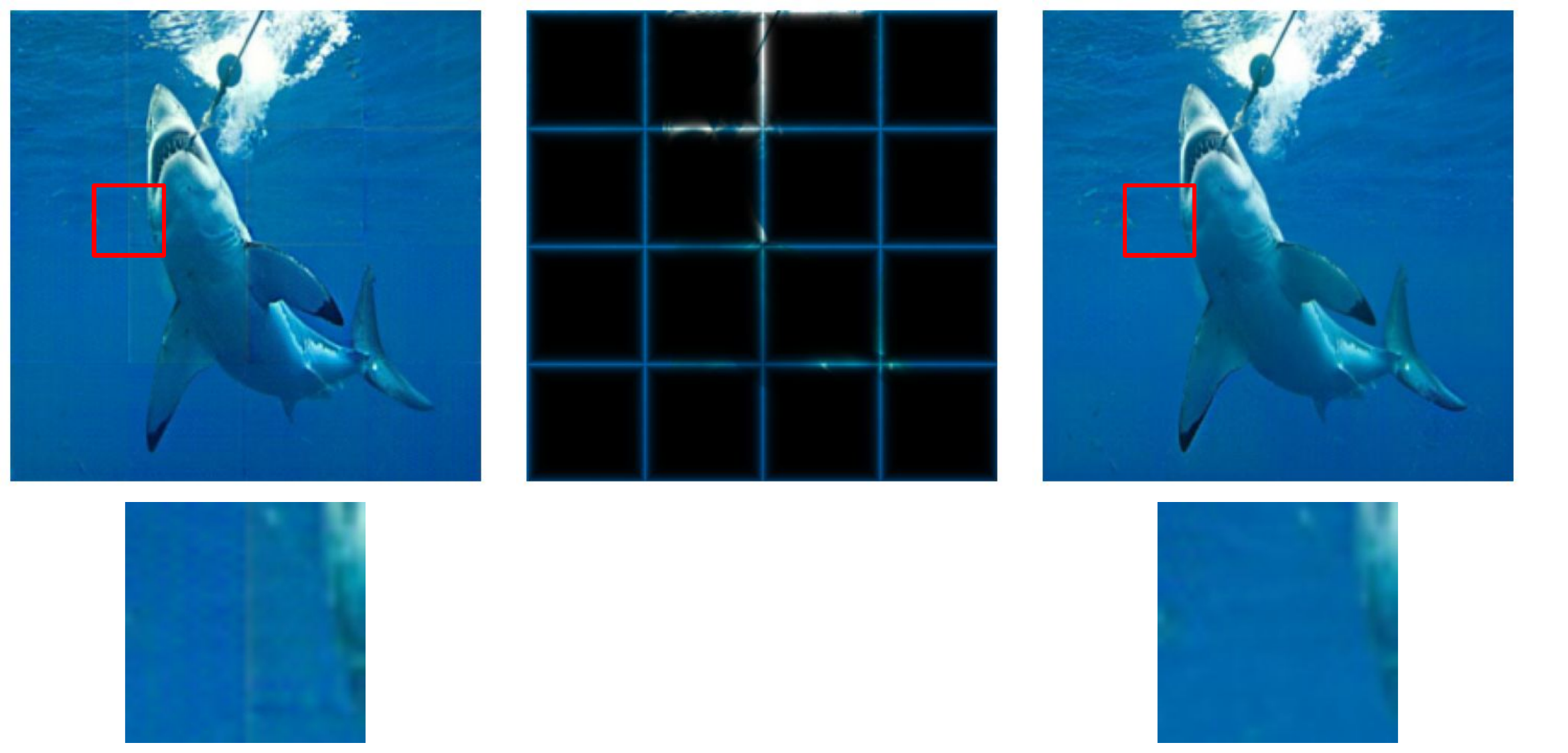}
    \vspace{-.5em}
    \caption{Blending the image to mitigate the visibility of Jigsaw edge artifacts.}
    \vspace{-1em}
    \label{fig:blendimg}
\end{figure}

\noindent\textbf{Minimizing Jigsaw Edge Visibility.} 
The Jigsaw process can leave watermarks with slightly visible effects despite not being reflected by similarity metrics like MSE, SSIM, or LPIPS (as we impose a strong image similarity loss during training). To enhance stealthiness, we create a mask ($M$) at the Jigsaw's segmenting edges (3-pixel width). This mask blends the original ($x$) and watermarked ($x_w$) images as $ x \cdot M + x_w \cdot (1-M) $. Fig.~\ref{fig:blendimg} demonstrates this blending, significantly improving watermark concealment to manual inspections.

\section{Experimental Settings Details}
\vspace{-.5em}
\subsection{Additional Details for $\AlgName$ Training}
\label{sec:details}
\vspace{-.5em}
\textbf{Detailed Random Perturbations.}
\addcontentsline{toc}{subsection}{\qquad Detailed Random Perturbations.}
\quad 
To effectively adapt to a variety of potential perturbations in real-world scenarios, our approach integrates a range of random perturbations during the contrastive learning process. This methodology is elaborated in Section \ref{sec:methodology}. Utilizing $\AlgName$, we bypass the need for gradient propagation through these perturbations and apply them in their unmodified, or 'vanilla', form. The perturbations we consider include JPEG compression, Gaussian blur, Gaussian noise, random rotations, brightness-contrast alterations, and the Diffusion-based image editing method, SDEdit, as referenced in \cite{meng2021sdedit}. Detailed parameters for these perturbations are listed in Table \ref{tab:detailperturbation}.

For each training image, we randomly select a combination of one to three of these perturbations, collectively referred to as $P$. The implementation details for each perturbation are as follows: For the mask, a random proportion of the image is obscured. For crop resize, a square section of the image is cropped and then resized back to the original dimensions. Random rotations involve either a horizontal or vertical flip, with the likelihood determined by a predefined probability. Lastly, for the SDEdit perturbation, we shuffle the editinginstructions and the image. This means each input image $x$ is modified using a random instruction from another image, enhancing model robustness and reducing the risk of overfitting.

The training process spans 100 epochs. We gradually increase the strength of the perturbations from a minimum to a maximum range, as outlined in Table \ref{tab:detailperturbation}. This increase follows a linear trajectory over the course of the training period.

\begin{table}[h]
\centering
\vspace{-.5em}
\resizebox{.95\linewidth}{!}{
\begin{tabular}{llll}
\textbf{Perturbation}                                                              & \textbf{Parameters}         & \textbf{Minimum  range} & \textbf{Maximum range} \\ \midrule[1.3pt]
\multirow{3}{*}{SDEdit~\cite{meng2021sdedit}}                                                   & Inference steps    & 50-80          & 50-80         \\
                                                                          & Strength           & 0.1-0.2        & 0.2-0.8       \\
                                                                          & Guidance scale     & 5-20           & 5-20          \\ \hline
JPEG                                                                      & Quality factor     & 10-30          & 20-90         \\ \hline
Mask                                                                      & Mask size          & 25-65          & 80-200        \\ \hline
Crop Resize                                                               & Crop Ratio         & 0.9-0.8        & 0.7-0.3       \\ \hline
\begin{tabular}[c]{@{}l@{}}Random\\ Rotate\end{tabular}                   & Rotate Probability & 0.5            & 0.5           \\ \hline
\begin{tabular}[c]{@{}l@{}}Contrast\\ Adjustment\end{tabular}             & Factor             & 0.16-0.3       & 0.8-1.5       \\ \hline
\begin{tabular}[c]{@{}l@{}}Brightness\\ Adjustment\end{tabular}           & Value              & 0-0.1          & 0-0.25        \\ \hline
\multirow{2}{*}{\begin{tabular}[c]{@{}l@{}}Guassian\\ Blur\end{tabular}}  & Kernel size        & 7              & 7             \\
                                                                          & Sigma              & 0.1-0.5        & 0.3-1.5       \\ \hline
\multirow{2}{*}{\begin{tabular}[c]{@{}l@{}}Guassian\\ Noise\end{tabular}} & Mean               & 0              & 0             \\
                                                                          & Standard deviation & 0.01-0.05      & 0.05-0.15    
\end{tabular}}
\caption{Detailed perturbation settings in the $\AlgName$ training.}
\vspace{-1em}
\label{tab:detailperturbation}
\end{table}

\noindent\textbf{$\AlgName$ Training Hypermeters.}
\addcontentsline{toc}{subsection}{\qquad Training Hypermeters.}
During the training of our model, we fine-tune the hyperparameters for both encoder and decoder, detailed in Table~\ref{tab:hyperparameters}. We utilized the AdamW~\cite{loshchilov2017decoupled} optimizer for its effectiveness in complex models. To regulate the training process, we applied a weight decay and momentum based on standard practices. The batch size and learning rate schedule were chosen to ensure both computational efficiency and steady convergence, with a warmup period easing the model into the full training regimen. These parameters are pivotal for achieving the desired optimization and generalization of our model.
\begin{table}[ht]
    \label{Tab}
    \begin{subtable}{.5\linewidth}
      \centering
        \resizebox{.95\linewidth}{!}{
        \begin{tabular}{l|l}
            \textbf{Config}             & \textbf{Value}          \\ \hline
            Optimizer          & AdamW          \\
            Base learning rate & 1e-4           \\
            Weight Decay       & 0.02           \\
            Momentum           & $\beta$1,$\beta$2 = 0.9,0.95 \\
            Batch Size         & 256            \\
            LR Schedule        & Cosine Decay   \\
            Warmup Epochs      & 10             \\
            Training Epochs    & 100           
        \end{tabular}
        }
        \caption{$\AlgName$ Encoder. $E$}
    \end{subtable}%
    \begin{subtable}{.5\linewidth}
      \centering
        \resizebox{.95\linewidth}{!}{
        \begin{tabular}{l|l}
            \textbf{Config}             & \textbf{Value}          \\ \hline
            Optimizer          & AdamW          \\
            Base learning rate & 2e-4           \\
            Weight Decay       & 0.05           \\
            Momentum           & $\beta$1,$\beta$2 = 0.9,0.95 \\
            Batch Size         & 768            \\
            LR Schedule        & Cosine Decay   \\
            Warmup Epochs      & 10             \\
            Training Epochs    & 100           
        \end{tabular}
        }
        \caption{$\AlgName$ Decoder, $D$}
    \end{subtable} 
    \vspace{-.5em}
    \caption{Hyperparameters for $\AlgName$ training.}
    \vspace{-1em}
\label{tab:hyperparameters}
\end{table}

\subsection{Evaluation Settings Details}
In this section, we will report the setting and hyperparameters that we use in each type evaluation in main paper.

\noindent\textbf{Type 1 - Conventional Perturbations.} 
\addcontentsline{toc}{subsection}{\qquad Type 1 - Conventional Perturbations.} 
Table~\ref{tab:type1eva} outlines the parameters used for conventional image perturbations. This table details the specific manipulations applied to assess the robustness of watermarked images under common transformations. It includes JPEG compression, which simulates the effects of lossy compression with a quality factor of 90, potentially introducing compression artifacts that could disrupt the watermark. Random rotation is tested with a 50\% probability, challenging the watermark's resilience to orientation changes. Contrast and brightness adjustments are evaluated with specific alteration levels to examine the watermark's stability under varying lighting conditions. Additionally, Gaussian Blur and Gaussian Noise are applied with defined kernel size, sigma, and standard deviation parameters to mimic the effects of blurring and noise – common artifacts in digital imaging. These perturbations and their hyperparameters are selected to represent real-world scenarios where watermarked images might be altered, which many are akin to the evaluation settings of existing watermark efforts~\cite{4554423, zhu2018hidden, fernandez2023stable, wen2023tree}.

\begin{table}[h]
\centering
\vspace{-.5em}
\resizebox{.6\linewidth}{!}{
\begin{tabular}{lll}
\textbf{Perturbation}                    & \textbf{Parameters}         & \textbf{Value} \\ \midrule[1.3pt]
JPEG                            & Quality factor     & 90    \\ \hline
Random Rotate                   & Rotate Probability & 0.5   \\ \hline
Contrast Adjustment             & Factor             & 1.0   \\ \hline
Brightness Adjustment           & Value              & 0.2   \\ \hline
\multirow{2}{*}{Guassian Blur}  & Kernel size        & 5     \\
                                & Sigma              & 0.3   \\ \hline
\multirow{2}{*}{Guassian Noise} & Mean               & 0     \\
                                & Standard deviation & 0.03 
\end{tabular}}
\caption{Hyperparameters of \textbf{Type 1} perturbations.}
\vspace{-1.em}
\label{tab:type1eva}
\end{table}

\noindent\textbf{Type 2 - Diffusion Perturbations.} 
\addcontentsline{toc}{subsection}{\qquad Type 2 - Diffusion Perturbations.} 
Table~\ref{tab:type2eva} presents the parameters for diffusion-based image perturbations. 
Akin to our threat model listed in Section \ref{sec:threatmodels}, we consider a list of unseen diffusion-based perturbations beyond the SDEdit (we incorporated in the training phase).
Similar to SDEdit, InstrucPix2Pix involves altering images through stochastic differential equations and text-to-image transformations, with varying levels of editingstrength, text guidance scale, and image guidance scale. The Zero 1-to-3 introduces diffusion-based perturbation that alters the viewpoint of images. InPaint evaluates the watermark's robustness against content-aware fill operations that significantly modify image content. Lastly, the impact of image variation via commercialized model DALL·E 2 on watermarks' detectablity is assessed. 
%
To synchronize our evaluations with the Human Aligned Variation ($\SimName$) scores ranging from 0.3 to 0.5 (discussed in Section \ref{sec:case1exp}), we adopt the $\SimName$ score as a filter during the image variation generation step leveraging different generative models. In particular, for each sample in the evaluation set (a total of 2000 samples), we iterative query the model with the sample, the paired instruction, and the hyperparameters listed in Table \ref{tab:type2eva} until a sample's $\SimName$ fallen into the range of 0.3-0.5. Note that only the perturbation from SDEdit is applied to training samples in our training phase of the $\AlgName$.


\begin{table}[h]
\centering
\vspace{-.5em}
\resizebox{0.7\linewidth}{!}{
\begin{tabular}{lcll}  
\textbf{Perturbation}                    & \textbf{Training Included} & \textbf{Parameters}           & \textbf{Value}    \\ \midrule[1.3pt]
\multirow{3}{*}{SDEdit}         & \multirow{3}{*}{Yes}                 & Inference Steps      & 50       \\
                                &                  & Modify Strength      & 0.3-0.6  \\
                                &                  & Text Guidance Scale  & 7.5-15   \\ \hline
\multirow{3}{*}{InstrucPix2Pix} & \multirow{3}{*}{No}                  & Inference Steps      & 50       \\
                                &                   & Text Guidance Scale  & 7.5-15   \\
                                &                   & Image Guidance Scale & 1.5-3    \\ \hline
\multirow{3}{*}{Zero 1-to-3}    & \multirow{3}{*}{No}                 & Inference Steps      & 50       \\
                                &                  & Polar angle          & -10 - 10 \\
                                &                  & Azimuth angle        & -10 - 10 \\ \hline
\multirow{2}{*}{InPaint}        & \multirow{2}{*}{No}                  & Inference Steps      & 50       \\
                                &                   & Mask Area            & 0.5-0.8  \\ \hline
DALL·E 2                        & No                 & NA                   & NA      
\end{tabular}}
\caption{Hyperparameters of \textbf{Type-2} perturbations.}
\vspace{-2em}
\label{tab:type2eva}
\end{table}

\noindent\textbf{Type 3 - Watermark Removal Attacks.} 
\addcontentsline{toc}{subsection}{\qquad Type 3 - Watermark Removal Attacks.} 
Table~\ref{tab:type3eva} delineates the parameters for various watermark removal attacks we considered in this paper. RG (ReGenerate)~\cite{zhao2023invisible} employs diffusion model to regenerate the original image and remove the imperceptible watermark. WEvade-B-Q\cite{jiang2023evading} focuses the attack on the decoder, using JPEG to heavily distort a watermarked image to erase the watermark, and employs HopSkipJump for black-box optimization to minimize perturbations by querying the decoder. Other baseline methods such as AdvH (Hihg Budget watermark adversarial attack)~\cite{saberi2023robustness}, attack the decoder through the transferability of adversarial examples. AC (Adversial Compression)~\cite{lukas2023leveraging} leverages an autoencoder to adversarially remove watermarks. All the above methods are adopted in a black-box setting, meaning they cannot directly access the decoder's gradient and parameters. To demonstrate watermark robustness under a white-box setting, we also include a PGD (Projected Gradient Descent)~\cite{madry2017towards} attack. It is noteworthy that such a PGD-based attack is also considered in WEvade~\cite{jiang2023evading} and AC~\cite{lukas2023leveraging} as their strongest settings of attack.

\begin{table}[h]
\centering
\vspace{-.5em}
\resizebox{.65\linewidth}{!}{
\begin{tabular}{llll}
\textbf{Perturbation}          & \textbf{Attack Type}      & \textbf{Parameters}      & \textbf{Value}     \\ \midrule[1.3pt]
RG                    & Black-box              & Inference Steps & 50        \\ \hline
\multirow{2}{*}{WEv}  & \multirow{2}{*}{Black-box}        & Query Times     & 100       \\
                      &         & $l_{\infty}$           & 8/255     \\ \hline
\multirow{3}{*}{AdvH} & \multirow{3}{*}{Black-box}        & Model           & ResNet-18 \\
                      &         & $l_{\infty}$           & 8/255     \\
                      &         & PGD Steps       & 40        \\ \hline
AC                    & Black-box              & Attack Iters    & 1         \\ \hline
\multirow{2}{*}{PGD}  & \multirow{2}{*}{White-box}        & $l_{\infty}$           & 8/255     \\
                      &         & PGD Steps       & 40       
\end{tabular}}
\caption{Hyperparameters of \textbf{Type 3} perturbations.}
\vspace{-2em}
\label{tab:type3eva}
\end{table}

\subsection{Dataset Settings Details}
\label{sec:modified_dataset}

\begin{table*}[h]
\centering
\resizebox{1.0\textwidth}{!}{
\begin{tabularx}{\textwidth}{lXl}
\toprule
\textbf{Category} & \textbf{Original Description by LLaVA} & \textbf{Edit Instruction by ChatGPT} \\
\midrule
Object Change & The image features a close-up of a large crab... & Add another crab. \\
& The image features a close-up of a large, green lizard... & Remove the lizard. \\
& The image features a woman sitting on the grass... & Replace the dog with a cat. \\
& The image shows a person holding a black bag... & Change the black bag to a red bag. \\
& The image features a dog standing on a wooden floor... & Make the dog run. \\
\midrule
Background Change & The picture features a mailbox sitting in a field... & Make the sky start raining. \\
& The image is a nighttime scene featuring a fish... & Replace the ground with a table. \\
& The picture features a man holding a black and white accordion... & Change the background color to green. \\
\midrule
Style Change & The image features a man wearing a hat... & Turn it into an oil painting style. \\
& The image features a man riding a motorcycle... & Change the helmet's material to metal. \\
\bottomrule
\end{tabularx}}
\caption{Various examples are given to ChatGPT to generate random Edit Instructions.}
\vspace{-1em}
\label{tab:modifyprompts}
\end{table*}

\begin{figure}[t]
    \centering
    \vspace{-.5em}
    \includegraphics[width=0.82\linewidth]{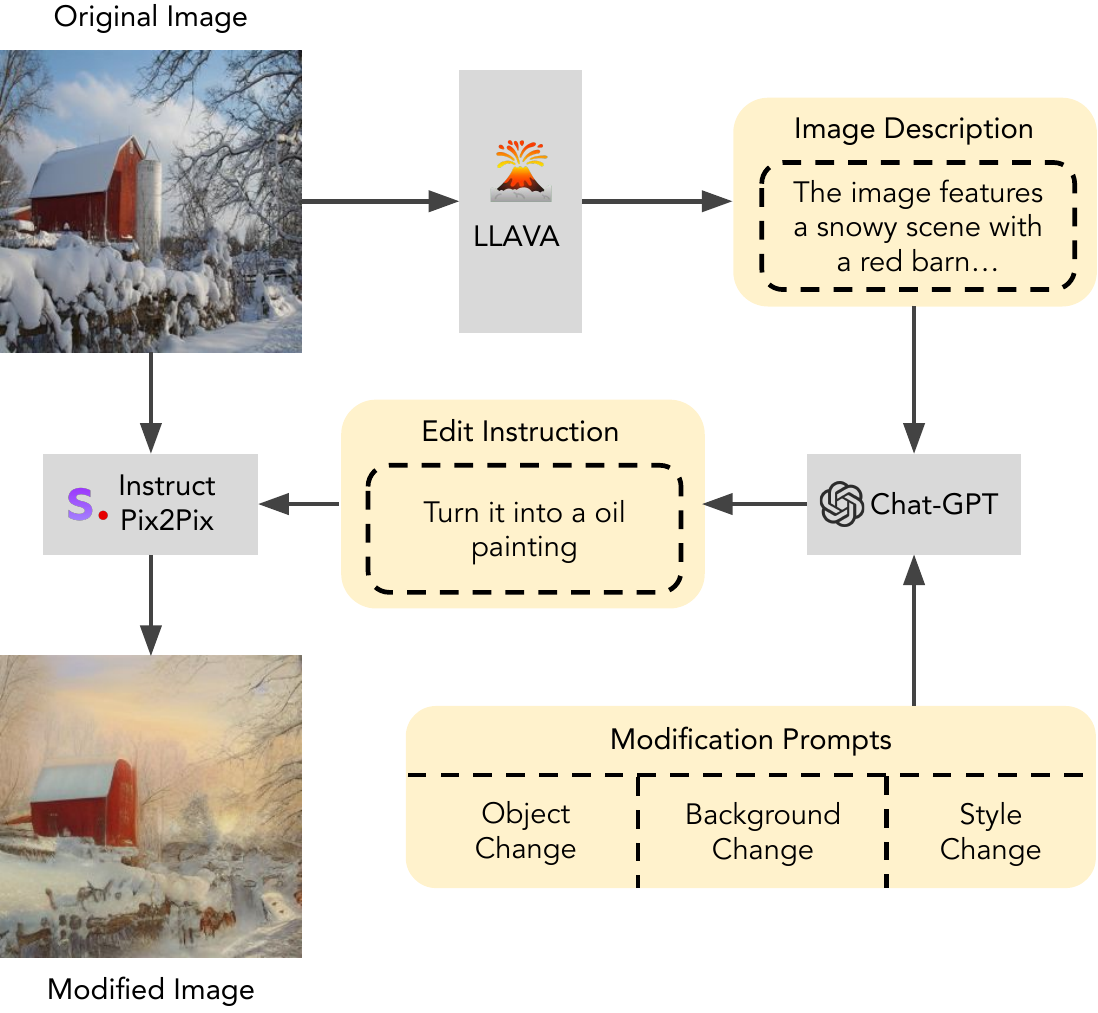}
    \vspace{-1em}
    \caption{Workflow for the creation and implementation of the image edit instructions dataset based on the ImageNet.}
    \label{fig:instrcdataset}
\end{figure}

In this section, we further detail the procedure of experiment set-up and how we adapt the ImageNet~\cite{5206848} dataset with corresponding edit instructions for our evaluation. As highlighted in Section~\ref{sec:mainevaluation}, due to potential data leakage issues that could impact the integrity of our results, the LAION-5B (which serve as part of the training set for all the considered diffusion models in this paper) and related datasets are excluded from the primary analysis. Instead, as ImageNet is not commonly used for image-text paired training of diffusion models and existing work had explored its' discrepancy to LAION-5B \cite{shirali2023makes}, we decided to proceed with the ImageNet and newly generated instructions by ourselves. However, to facilitate a comprehensive assessment, evaluations utilizing the LAION-5B dataset are included and discussed in Section~\ref{append:diffdataset}.

For the generation of ImageNet evaluation dataset, a comprehensive illustration of this process is provided in Fig.~\ref{fig:instrcdataset}. The procedure begins with the original images from the ImageNet, which is fed into LLaVA~\cite{liu2023visual}, a visual-language model. This model can respond to textual queries based on the provided image. By prompting the question ``What is the content of the image?'' to LLaVA, it yields an approximate 60-token-sized description of the image. This description, along with the editingprompts from Table~\ref{tab:modifyprompts}, is then input into ChatGPT~\cite{ouyang2022training}. ChatGPT is then prompt to generate editing instructions based on this input. These image description and instructions will subsequently fed into our diffusion-based image editingtool, such as SDEdit~\cite{meng2021sdedit} and InstructPix2Pix~\cite{brooks2023instructpix2pix}, to produce the final modified image variation results. 

\noindent\textbf{Training Dataset Settings for $\AlgName$.}
For the training dataset, we employ the previously mentioned method to generate editing instructions for the ImageNet~\cite{5206848} test dataset, which contains 100,000 images across 1,000 different classes. During the training of $\AlgName$, we shuffle the editing instructions for each image when loading the images to the SDEdit to simulate more drastic instructions.

\noindent\textbf{Evaluation Dataset Settings.}
For our evaluation dataset used in evaluating Type 2 perturbations, we apply the same method previously described for generating editing instructions, this time focusing on the ImageNet~\cite{5206848} validation dataset. This dataset encompasses 50,000 images across 1,000 distinct classes. We randomly select a subset of 2,000 samples to generate our evaluation dataset in Section \ref{sec:case2exp}. 


\section{Design Choice \& Engineering Details}
This section examines key aspects of our watermarking model, including the effectiveness of various loss functions, the impact of different model architectures, and essential design enhancements such as gradient clipping and Jigsaw edge visibility minimization, to ensure robust watermark detection and model optimization.
\subsection{Ablation for Loss Function}
\label{sec:ablation}
In this section, we evaluated the effectiveness of various loss functions for a binary classification task in watermark detection, as present in Table~\ref{tab:losstype}. Our analysis compared Mean Squared Error (MSE), Binary Cross-Entropy (BCE), Focal Loss, Balanced Discriminative (BD), and BD with a temperature threshold ($\tau=0.1$), focusing on their AUC performance. MSE showed limited effectiveness with an AUC of 0.675, likely due to its generic approach not specifically tailored for binary classification. BCE, better suited for such tasks, improved the AUC to 0.721. Focal Loss, addressing class imbalance by emphasizing hard-to-classify cases, further enhanced the AUC to 0.733. The BD loss, aiming for balanced training across positive and negative classes, achieved an AUC of 0.738.
The introduction of a temperature threshold ($\tau=0.1$) in the BD loss, known as Temperature Binomial Deviance Loss (TBDL), significantly improved performance, yielding the highest AUC of 0.781. This temperature parameter intensifies the model's focus on examples near the decision boundary, enhancing its sensitivity to difficult cases and boosting overall accuracy in distinguishing between watermarked and non-watermarked images. As a result of these insights, we have chosen the BD loss with a temperature threshold as the primary loss function for $\AlgName$ in our paper.

\begin{table}[h]
\centering
\vspace{-.5em}
\resizebox{.6\linewidth}{!}{
\begin{tabular}{c|ccccc}
\textbf{Loss Type} & MSE & BCE  & Focal  & BD  & BD$\tau=0.1$         \\ \hline
\textbf{AUC}       & 0.675    & 0.721 & 0.733      & 0.738      & \textbf{0.781}
\end{tabular}}
\caption{Different loss type and their result.}
\vspace{-3em}
\label{tab:losstype}
\end{table}

\subsection{Ablation of Encoder/Decoder Architecture}
\begin{figure}[!t]
\vspace{-.5em}
    \begin{center}
    \includegraphics[width=0.95\linewidth]{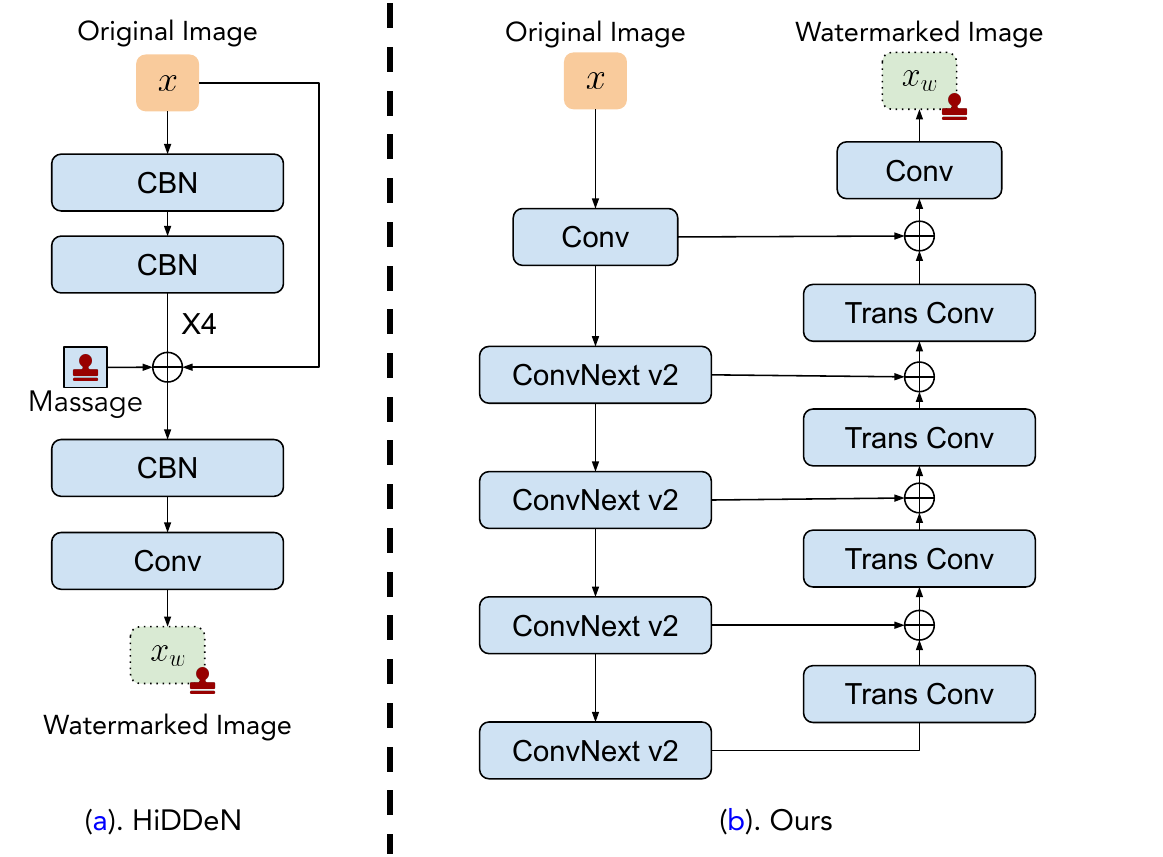}
    \end{center}
    \vspace{-1.em}
     \caption{$\AlgName$'s different encoder network structures comparing to existing work (different base block and connections).}
     \vspace{-1.5em}
    \label{fig:network}
\end{figure}

At the heart of the watermarking framework lies the watermarking model, which plays a pivotal role in determining the watermark quality and final detection performance. Consequently, the architecture of this model is of paramount importance. However, the designs of both the encoder and decoder have not been extensively explored in prior research. HiDDeN~\cite{zhu2018hidden} pioneered the encoder-decoder structure, wherein both components were constructed using multiple Conv-BN-ReLU (CBR) blocks, as shown in Fig.~\ref{fig:network}\textcolor{blue}{a}.

Building on HiDDeN's foundation, several studies have adopted this basic model structure~\cite{liu2019novel, fernandez2023stable}. Some advancements, like StegaStamps~\cite{tancik2020stegastamp}, have enhanced the model by substituting the basic encoder with a U-Net~\cite{ronneberger2015u}, which still employs CBR blocks. The U-Net architecture is specifically designed to capture hierarchical image information through its encoder-decoder structure. Recent advancements like MBRS~\cite{jia2021mbrs} have transitioned from CBR to SENet Blocks~\cite{hu2018squeeze} for both encoder and decoder, enabling the model to concentrate on crucial image regions. While these models exhibit commendable performance in their specific scenarios, their relatively simplistic and suboptimal designs are insufficient for tackling the watermarking challenges presented by $\AlgName$, as detailed in Table~\ref{tab:prenetfail}.

\begin{table}[h]
\centering
\vspace{-.5em}
\resizebox{.45\linewidth}{!}{
\begin{tabular}{cccc}
\textbf{Encoder} & \textbf{Decoder} & \textbf{PSNR}  & \textbf{AUC}   \\ \midrule[1pt]
CBR   & CBR   & \textbf{28.07} & 0.733 \\
UNet    & CBR   & 28.02 & \textbf{0.773} \\
SE      & SE      & 28.17 & 0.767
\end{tabular}}
\caption{Performance comparison for different architectures.}
\vspace{-1em}
\label{tab:prenetfail}
\end{table}
While there are evident performance differences, directly adopting these model designs might result in redundant architectures, particularly given the unique optimization objectives of $\AlgName$. To address this, we propose an exhaustive ablation study on various model designs. This will allow us to fully grasp the significance of each model component and rethink the architecture, ensuring we identify the most streamlined and effective solution for the diffusion watermarking challenge.

Before delving into the intricacies of model design, it's essential to understand the role of each component. The encoder's primary function is to seamlessly embed the watermark into the host image. In contrast, the decoder's role is to extract the watermark score from the watermarked image. In real-world applications, users may not always know which images contain watermarks. As a result, they might feed both watermarked and unwatermarked images into the decoder, which further causes more query time than the encoder. This underscores the decoder's twofold significance: ensuring robust detection and optimizing computational efficiency at inference.

\noindent\textbf{Remove Key Embedding Layers.} Since $\AlgName$ does not require a predefined watermark message as a watermark for input, we can eliminate both the secret key encoder and the concat layer that merges the watermark information with the image. 
The computational and performance results of these modifications are presented in Table~\ref{tab:redundant}. It is evident from the table that the message encoder and the concat layer contribute to increased computational complexity. By omitting these components, our method enhances both image quality and detection performance. We will keep this design in all the later experiments.
\begin{table}[h]
\centering
\vspace{-.5em}
\resizebox{.55\linewidth}{!}{
\begin{tabular}{cccc}
\textbf{Model Design }       & \textbf{Flops}  & \textbf{PSNR}  & \textbf{AUC}   \\ \midrule[1pt]
Raw                 & 11.12G & 28.07 & 0.733 \\
w/o message encoder & 9.99G  & 28.11 & 0.736 \\
w/o concat layer   & \textbf{7.44G}  & \textbf{28.24} & \textbf{0.738}
\end{tabular}}
\caption{Performance after removing key embedding layers.}
\vspace{-1em}
\label{tab:redundant}
\end{table}

\noindent\textbf{Encoder Depth \& Width Ablation.} 
Although the HiDDeN model is originally designed without down sampling layers, recent studies have shown that down sampling can reduce computational complexity and enhance model performance by capturing higher-level information from input features~\cite{chatterjee2022generalization}. Besides the depth of the model, the width, represented by the number of channels, also plays an important role. The original design uses 64 channels. 
To understand the impact of down sampling and channel width on performance, we conducted an ablation study, summarized in Table~\ref{tab:downsampling}. Our findings suggest that while introducing down sampling can lead to a reduction in image recovery performance, it significantly reduces computational complexity. On the other hand, increasing the number of inner channels positively impacts the model's performance. Models without down sampling exhibit a notable increase in computational complexity. Considering the trade-offs, we identified the 2x down sampling block with 128 channels as the optimal balance between complexity and performance. This configuration not only outperforms the original design in terms of PSNR and AUC but also achieves this with reduced computational overhead.
\begin{table}[h]
\centering
\vspace{-.5em}
\resizebox{.55\linewidth}{!}{
\begin{tabular}{ccccc}
\textbf{Down sampling} & \textbf{\#Channels} & \textbf{Flops} & \textbf{PSNR}  & \textbf{AUC}   \\ \midrule[1pt]
0x                     & 64                  & 7.44G         & 28.24          & 0.738          \\
0x                     & 128                 & 29.38G         & \textbf{31.36} & \textbf{0.747} \\
2x                     & 64                  & 1.86G          & 27.62          & 0.731          \\
2x                     & 128                 & 7.34G         & 31.23          & 0.743          \\
4x                     & 64                  & \textbf{0.46G}          & 26.82          & 0.712          \\
4x                     & 128                 & 1.84G           & 27.41          & 0.728         
\end{tabular}}
\caption{Performance with varying down sampling and channels.}
\vspace{-1em}
\label{tab:downsampling}
\end{table}

\noindent\textbf{Design Choice of Watermark Encoder.} 
In the StegaStamps~\cite{tancik2020stegastamp}, the watermark encoder employs a U-Net architecture~\cite{ronneberger2015u}. 
The U-Net architecture is characterized by its U-shaped structure, comprising a downsampling path and an upsampling path. Skip connections bridge the downsampling and upsampling paths, facilitating improved image reconstruction quality. This architecture enables the embedding of watermarks into both the high-level semantics and the intricate details of the image. To evaluate the impact of the U-Net's depth on our encoder model's performance, we implemented the U-Net structure with varying depths. The outcomes of this investigation are summarized in Table~\ref{tab:unetdepth}.
Our results suggest that deeper architectures enhance watermarking performance due to their ability to embed watermarks across diverse image levels. The increasing detection AUC with depth supports this observation. Although there's a minor trade-off in visual quality, the improvements in watermark detection justify this compromise. Based on these observations, we selected a U-Net configuration with 4 downsampling layers for subsequent experiments.
\begin{table}[h]
\centering
\vspace{-.5em}
\resizebox{.35\linewidth}{!}{
\begin{tabular}{ccc}
\textbf{UNet Depth} & \textbf{PSNR}                      & \textbf{AUC}                       \\ \midrule[1pt]
0     & 31.23                     & 0.743                     \\
1     & \textbf{31.36}                     & 0.762                     \\
2     & 31.28                     & 0.769                     \\
3     & 31.22 & 0.777 \\
4     & 31.12 & \textbf{0.783}
\end{tabular}}
\caption{Performance across varying U-Net depths.}
\vspace{-2em}
\label{tab:unetdepth}
\end{table}

\noindent\textbf{Encoder Basic Block Type Ablation.} 
The initial implementation in HiDDeN employs naive CBR blocks as the fundamental unit in both the encoder and decoder. The MBRS approach\cite{jia2021mbrs} enhances performance by replacing the CBR with Squeeze-and-Excitation Networks (SENet) Blocks\cite{hu2018squeeze}. However, with the rapid advancements in deep learning, various network architectures have been proposed to achieve state-of-the-art performance\cite{he2016deep, hu2018squeeze, tan2021efficientnetv2, woo2023convnext}. To understand the impact of different block types, we pick the most representative work to replace the original CBR block and show the results in Table~\ref{tab:basicblock}.
ConvNeXt V2, evolving from traditional convolutional architectures, uniquely combines depth convolutional design with a Global Response Normalization layer, enabling better performance in various recognition benchmarks \cite{woo2023convnext}. By adopting such blocks, we observed significant improvements in both visual quality and watermark detection performance.
\begin{table}[h]
\centering
\vspace{-.5em}
\resizebox{.35\linewidth}{!}{
\begin{tabular}{ccc}
\textbf{Block Type}   & \textbf{PSNR}                      & \textbf{AUC}                       \\ \midrule[1pt]
Conv-BN-ReLU & 28.12                     & 0.783                     \\
Residual~\cite{he2016deep}     & 29.64                     & 0.816                     \\
SE~\cite{hu2018squeeze}           & 30.34                     & 0.837                     \\
MBConv~\cite{tan2021efficientnetv2}       & 29.07 & 0.811 \\
ConvNeXt V2~\cite{woo2023convnext}     & \textbf{30.83} & \textbf{0.856}
\end{tabular}}
\caption{Comparison of different basic block types.}
\vspace{-2em}
\label{tab:basicblock}
\end{table}

\noindent\textbf{Decoder Backbone Model Ablation.} For the decoder, as we have already reformed the watermarking task into a binary classification task, any existing classification model can be adopted without limitation. On the other hand, recalling our aim for the decoder: less overhead and better detection performance, our focus is on lightweight and inference-efficient models. Luckily, such efficient models have been widely researched~\cite{howard2019searching, tan2021efficientnetv2, tan2019mnasnet, ma2018shufflenet}, and we can easily adopt any of them as the $\AlgName$ decoder. Table~\ref{tab:decodermodel} shows the results of some of the most representative models as detectors:
\begin{table}[h]
\centering
\vspace{-.5em}
\resizebox{.55\linewidth}{!}{
\begin{tabular}{cccc}
\textbf{Model}           & \textbf{Flops}          & \textbf{\#Param }      & \textbf{AUC}            \\ \midrule[1pt]
Plain           & 15.87G         & \textbf{2.4M} & 0.856          \\
MobileNetV3-L~\cite{howard2019searching}   & \textbf{0.22G} & 5.5M          & 0.943          \\
EfficientV2-S~\cite{tan2021efficientnetv2}   & 8.37G          & 21.5M         & \textbf{0.946} \\
MnasNet1-3~\cite{tan2019mnasnet}      & 0.53G          & 6.3M          & 0.928          \\
ShuffleNetV2-X2~\cite{ma2018shufflenet} & 0.58G          & 7.4M          & 0.937         
\end{tabular}}
\caption{Performance comparison of various decoder models.}
\vspace{-2em}
\label{tab:decodermodel}
\end{table}
Considering the trade-off between computational cost (Flops) and performance (AUC), we have selected MobileNetV3-L~\cite{howard2019searching} as our final detector structure due to its efficiency and competitive performance.

\vspace{-0.5em}
\section{Qualitative Study}
\vspace{-0.5em}
\subsection{$\SimName$ Visual Reflections}
\begin{figure*}[h]
    \centering
    \includegraphics[width=0.98\linewidth]{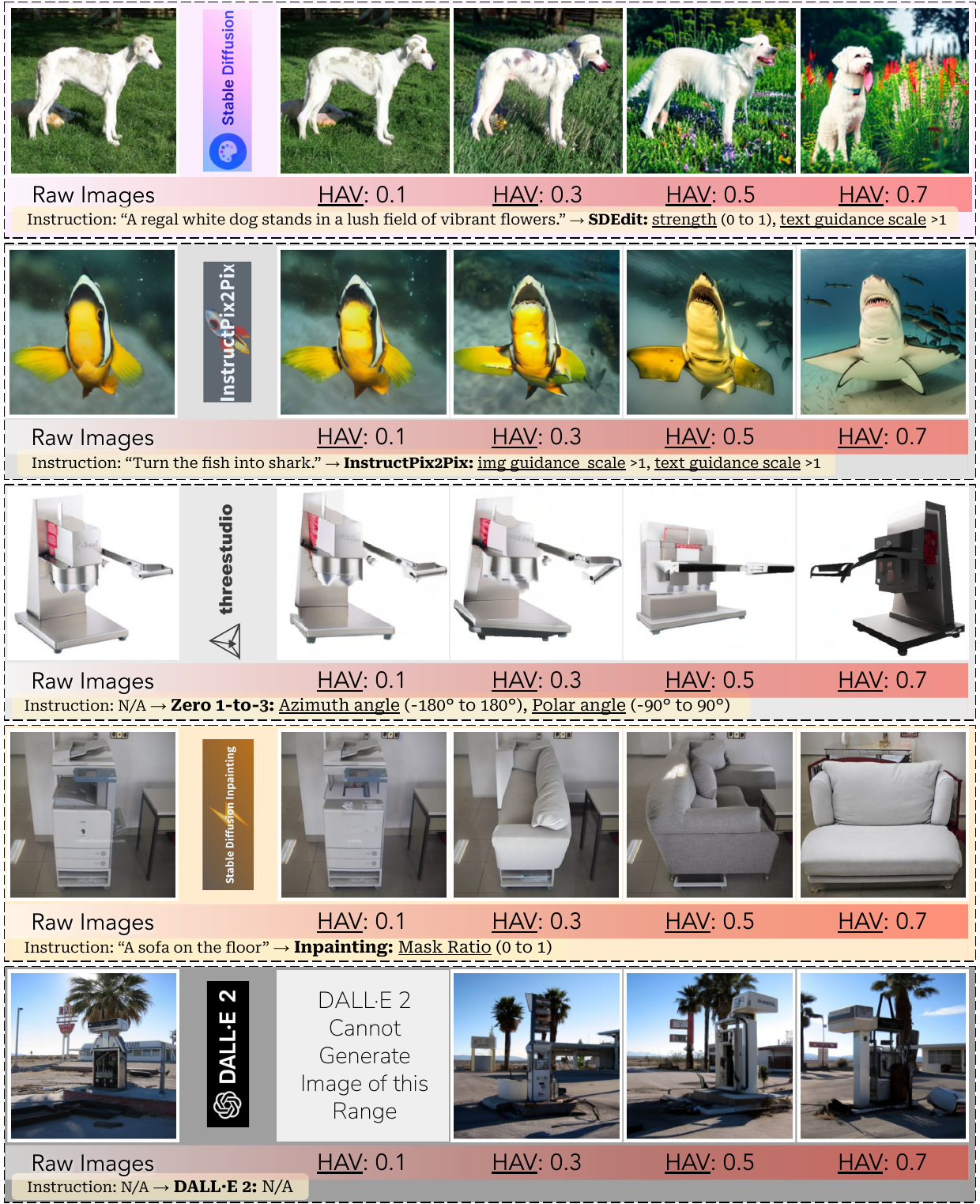}
    \vspace{-.5em}
    \caption{Visual representation of image modifications and corresponding $\SimName$ scores. This figure showcases a series of images with varying degrees of modifications via different methods, each annotated with its relative $\SimName$ score. Additionally, we include the specific hyperparameters and instructions employed for each modification.
    }
    \vspace{-1em}
    \label{fig:hav_visual}
\end{figure*}

\vspace{-.5em}
Fig.~\ref{fig:hav_visual} presents a series of visual examples to demonstrate the capability of $\SimName$ in evaluating image modifications. The figure pairs various altered images with their corresponding $\SimName$ scores, exemplifying the metric's alignment with human judgment across a spectrum of alteration techniques and their parameters. It encapsulates the diversity of editingmechanisms—such as viewpoint adjustments or object changes—and highlights the unique control each method offers over the alteration extent, with the exception of DALL$\cdot$E 2 where such control is not user-determined. Despite the various kinds of changes, $\SimName$ reliably indicates an aligned score to human perception of information derivative.

\vspace{-.5em}
\subsection{$\AlgName$ Visual Qualities}
\begin{figure*}[h]
    \centering
    \includegraphics[width=0.98\linewidth]{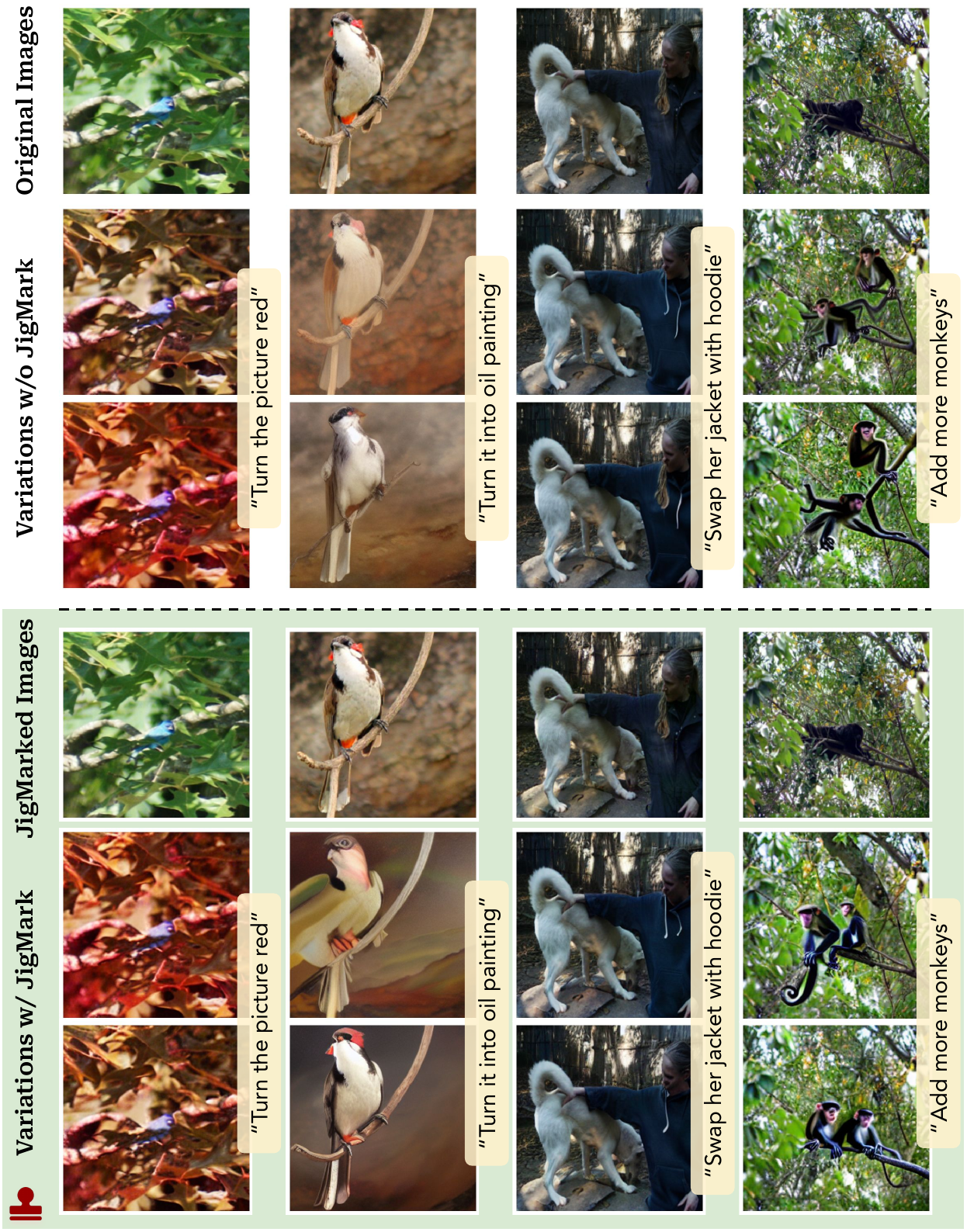}
    \vspace{-.5em}
    \caption{
    Visualization of original and $\AlgName$ processed images under InstructPix2pix~\cite{brooks2023instructpix2pix} perturbation. We randomly select two images for their diffusion-perturbed view to demonstrate the various effects of diffusion perturbation. Notably, $\AlgName$ preserves the visual quality effectively across both the original and perturbed images, showcasing its stealthiness.}
    \vspace{-1em}
    \label{fig:jigmark_visual}
\end{figure*}

\vspace{-.5em}
To illustrate the visual impact and stealthiness of $\AlgName$, we randomly select two samples for modification, with the resulting visualizations presented in Fig.~\ref{fig:jigmark_visual}. As evident in the samples, applying $\AlgName$ leaves not much perceptible trace on the original images themselves nor influences the quality of the diffusion generative process.

\end{document}